%% file: main.tex
\documentclass[lettersize,journal]{IEEEtran}
\usepackage{lipsum} % for mock text
\usepackage{amsmath,amsfonts}
\usepackage{algorithmic}
\usepackage{algorithm}
\usepackage{array}
\usepackage[caption=false,font=normalsize,labelfont=sf,textfont=sf]{subfig}
\usepackage{textcomp}
\usepackage{stfloats}
\usepackage{url}
\usepackage{verbatim}
\usepackage{graphicx}
\usepackage{cite}
\usepackage{comment}
\usepackage{multirow}
\usepackage{xcolor}
\usepackage{amssymb}
\usepackage{booktabs}
\usepackage[hidelinks]{hyperref}
\usepackage{etoolbox,xspace}
\usepackage{cuted}
\usepackage{capt-of}
\usepackage{enumitem}

\def\shownotes{0}

\ifnum\shownotes=1
\newcommand\zhiming[1]{\textcolor{cyan}{Zhiming: #1}}
\newcommand\hl[1]{\textcolor{red}{#1}}
\newcommand\syn[1]{\textcolor{green}{Syn: #1}}
\newcommand\todo[1]{\textcolor{red}{#1}}
\newcommand\andreas[1]{\textcolor{orange}{Andreas: #1}}
\else 
\newcommand\andreas[1]{}
\newcommand\syn[1]{}
\newcommand\zhiming[1]{}
\newcommand\hl[1]{#1}
\newcommand\todo[1]{}
\fi

\usepackage{graphicx}
\usepackage{etoolbox}

\begin{document}
\newcommand{\methodName}{Pose2Gaze\xspace}
\title{\methodName: Eye-body Coordination during Daily Activities for Gaze Prediction from Full-body Poses}

\author{Zhiming Hu, Jiahui Xu, Syn Schmitt, Andreas Bulling \IEEEcompsocitemizethanks{\IEEEcompsocthanksitem 
Zhiming Hu, Jiahui Xu, Syn Schmitt, and Andreas Bulling are with the University of Stuttgart, Germany. \protect E-mail: \{zhiming.hu@vis.uni-stuttgart.de, st170522@stud.uni-stuttgart.de, schmitt@simtech.uni-stuttgart.de, andreas.bulling@vis.uni-stuttgart.de\}. Syn Schmitt and Andreas Bulling are with the Center for Bionic Intelligence Tuebingen Stuttgart (BITS), Germany. Zhiming Hu is the corresponding author.
}

\thanks{Manuscript received January 26, 2024.}}

%\markboth{IEEE Transactions on Visualization and Computer Graphics, January~2024}%


\maketitle
\input{sections/abstract}
\input{sections/introduction}
\input{sections/related_work}
\input{sections/analysis}
\input{sections/method}
\input{sections/experiments}

\input{sections/application}
\input{sections/discussion}
\input{sections/conclusion}

\section*{Acknowledgement}
This work was funded by the Deutsche Forschungsgemeinschaft (DFG, German Research Foundation) under Germany's Excellence Strategy -- EXC 2075 -- 390740016.
A. Bulling was funded by the European Research Council (ERC) under grant agreement 801708.

{
    %\small
    \bibliographystyle{IEEEtran}
    \bibliography{references.bib}
}

%\newpage

\section{Biography Section}
%If you have an EPS/PDF photo (graphicx package needed), extra braces are needed around the contents of the optional argument to biography to prevent the LaTeX parser from getting confused when it sees the complicated $\backslash${\tt{includegraphics}} command within an optional argument. (You can create your own custom macro containing the $\backslash${\tt{includegraphics}} command to make things simpler here.)
\vspace{-30pt}

\begin{IEEEbiography}[{\includegraphics[width=1in,height=1.25in,clip,keepaspectratio]{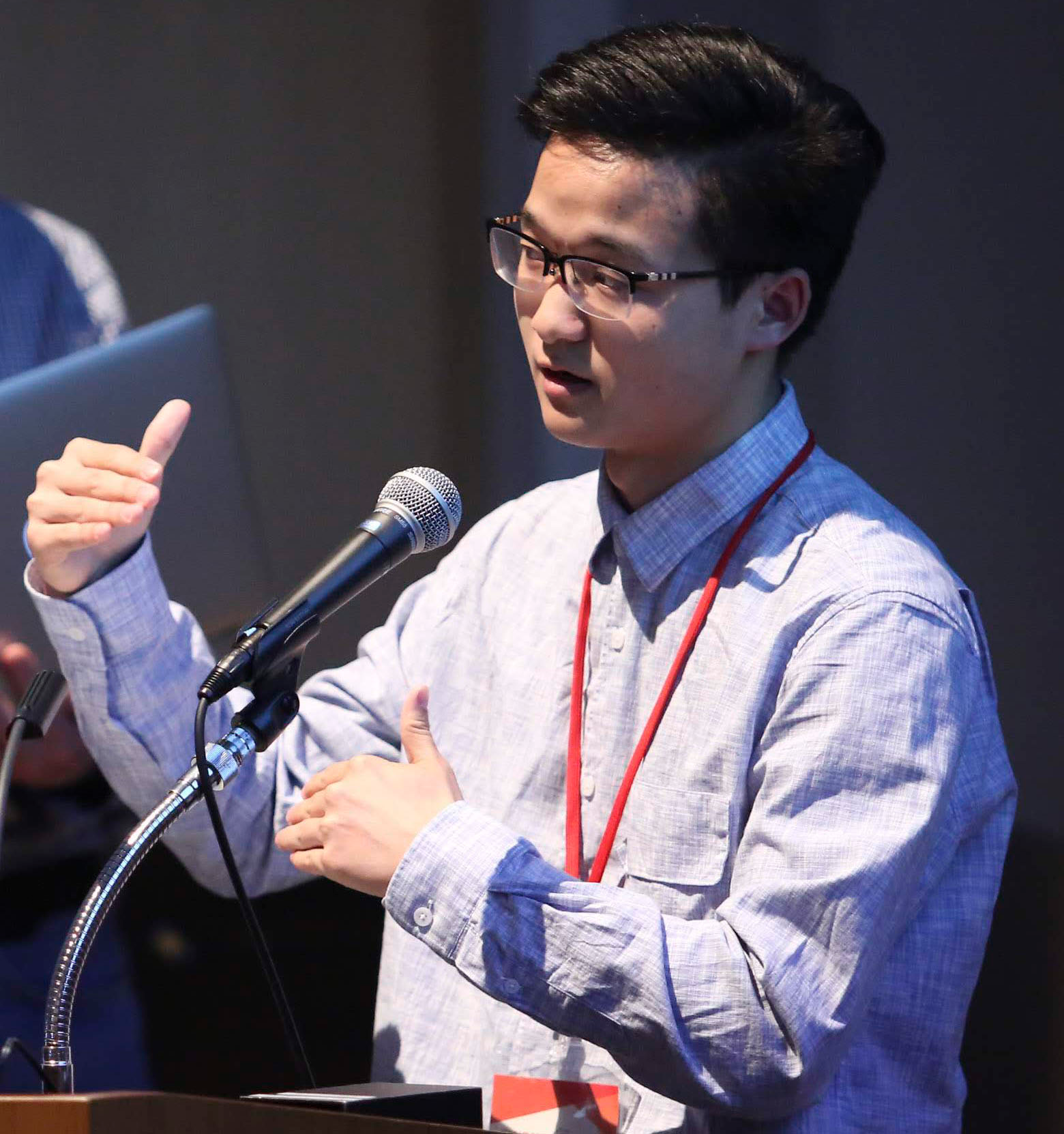}}]{Zhiming~Hu} 
is a post-doctoral researcher at the University of Stuttgart, Germany. 
He obtained his Ph.D. degree in Computer Software and Theory from Peking University, China in 2022 and received his Bachelor's degree in Optical Engineering from Beijing Institute of Technology, China in 2017. His research interests include virtual reality, human-computer interaction, eye tracking, and human-centred artificial intelligence.
\end{IEEEbiography}

\vspace{-35pt}

\begin{IEEEbiography}[{\includegraphics[width=1in,height=1.25in,clip,keepaspectratio]{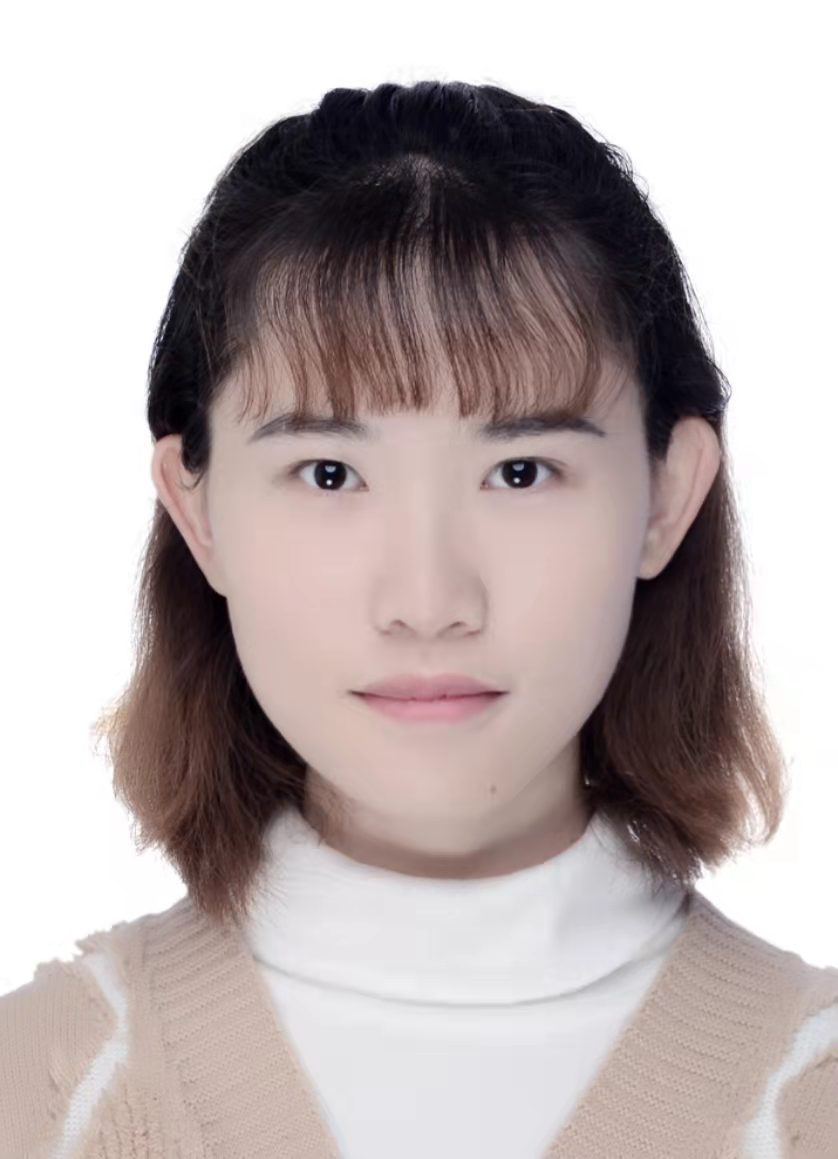}}]{Jiahui~Xu}
is a master student at the University of Stuttgart, Germany. She received her Bachelor's degree in Electronic Information Engineering from Jilin University, China in 2019. Her research interests include human-computer interaction, computer vision, and virtual reality.
\end{IEEEbiography}

\vspace{-35pt}

\begin{IEEEbiography}[{\includegraphics[width=1in,height=1.25in,clip,keepaspectratio]{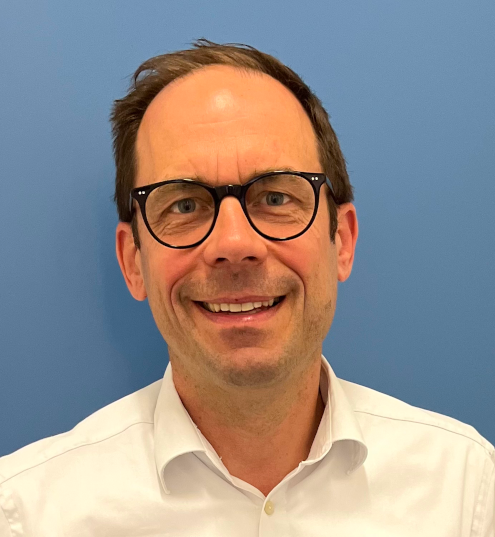}}]{Syn~Schmitt}
is Full Professor of Biomechanics and Biorobotics at the University of Stuttgart, Germany, where he directs the research group "Computational Biophysics and Biorobotics". He received his MSc. in Physics from the University of Stuttgart, Germany, in 2003 and his PhD in Theoretical Physics from University of Tübingen, Germany, in 2006. His research interests include biomechanics, neuromechanics, biorobotics.
\end{IEEEbiography}

\vspace{-35pt}

\begin{IEEEbiography}[{\includegraphics[width=1in,height=1.25in,clip,keepaspectratio]{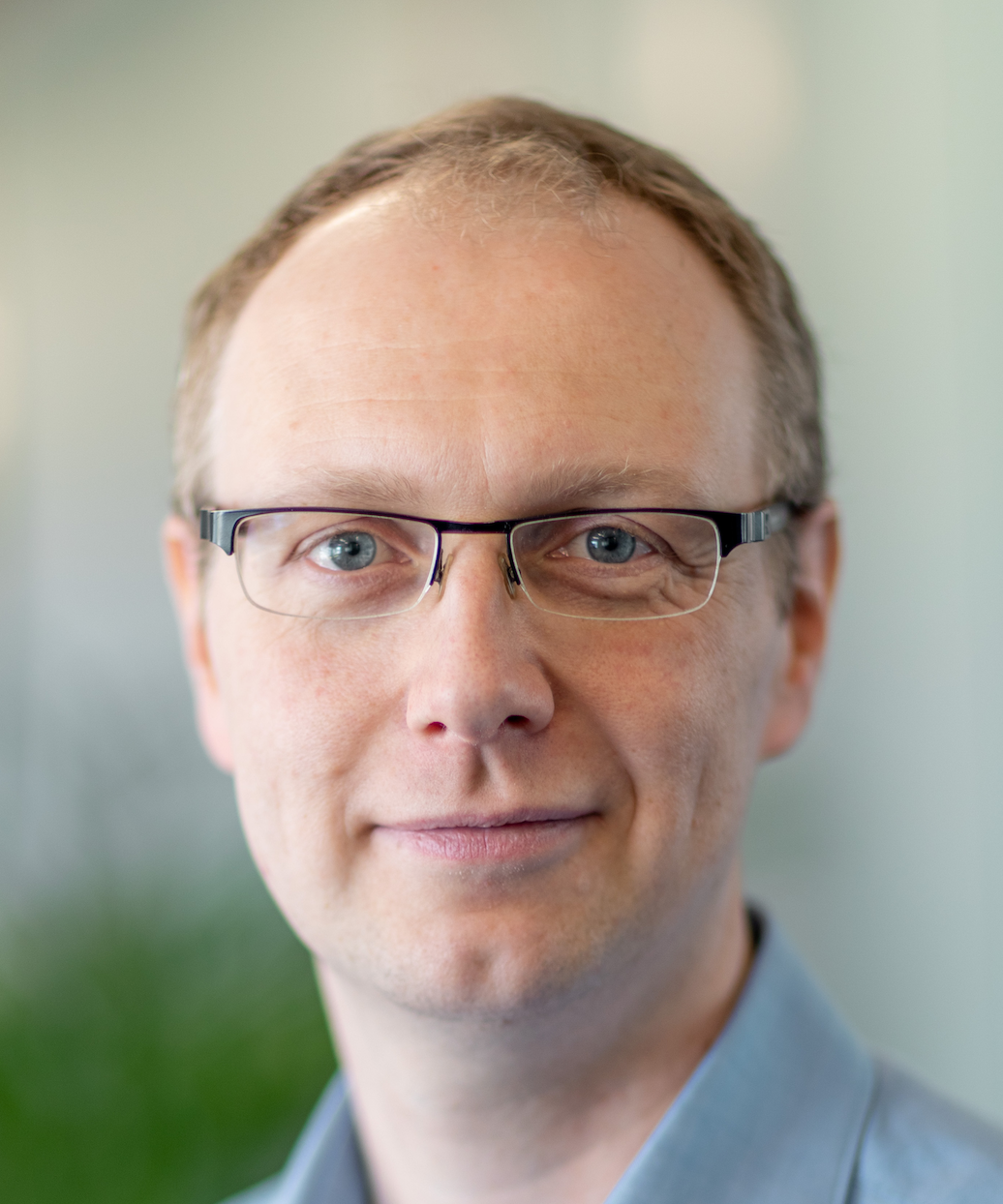}}]{Andreas~Bulling}
is Full Professor of Computer Science at the University of Stuttgart, Germany, where he directs the research group "Human-Computer Interaction and Cognitive Systems". He received his MSc. in Computer Science from the Karlsruhe Institute of Technology, Germany, in 2006 and his PhD in Information Technology and Electrical Engineering from ETH Zurich, Switzerland, in 2010. Before, Andreas Bulling was a Feodor Lynen and Marie Curie Research Fellow at the University of Cambridge, UK, and a Senior Researcher at the Max Planck Institute for Informatics, Germany. His research interests include computer vision, machine learning, and human-computer interaction.
\end{IEEEbiography}

\vfill
\end{document}

%% file: sections/abstract.tex
%%%%%%%%% ABSTRACT

\begin{abstract}
Human eye gaze plays a significant role in many virtual and augmented reality (VR/AR) applications, such as gaze-contingent rendering, gaze-based interaction, or eye-based activity recognition.
However, prior works on gaze analysis and prediction have only explored eye-head coordination and were limited to human-object interactions.
We first report a comprehensive analysis of eye-body coordination in various human-object and human-human interaction activities based on four public datasets collected in real-world (MoGaze), VR (ADT), as well as AR (GIMO and EgoBody) environments.
We show that in human-object interactions, e.g. \textit{pick} and \textit{place}, eye gaze exhibits strong correlations with full-body motion while in human-human interactions, e.g. \textit{chat} and \textit{teach}, a person's gaze direction is correlated with the body orientation towards the interaction partner.
Informed by these analyses we then present \textit{Pose2Gaze}~-- a novel eye-body coordination model that uses a convolutional neural network and a spatio-temporal graph convolutional neural network to extract features from head direction and full-body poses, respectively, and then uses a convolutional neural network to predict eye gaze.
We compare our method with state-of-the-art methods that predict eye gaze only from head movements and show that \textit{Pose2Gaze} outperforms these baselines with an average improvement of $24.0\%$ on MoGaze, $10.1\%$ on ADT, $21.3\%$ on GIMO, and $28.6\%$ on EgoBody in mean angular error, respectively.
We also show that our method significantly outperforms prior methods in the sample downstream task of eye-based activity recognition.
These results underline the significant information content available in eye-body coordination during daily activities and open up a new direction for gaze prediction.
\end{abstract}

\begin{IEEEkeywords}
Eye-body coordination, human-object interaction, human-human interaction, gaze prediction, activity recognition, virtual reality, augmented reality
\end{IEEEkeywords}

%% file: sections/introduction.tex
\section{Introduction}
% The importance of eye gaze
With the increasing use of virtual and augmented reality (VR/AR), understanding and predicting human behaviour in VR/AR environments has become a popular research topic and is a key requirement for intelligent human-aware VR/AR agents~\cite{hu2021fixationnet, hadnett2019effect}.
Human eye movements are of particular interest given that they are key for important VR/AR applications, such as 1) gaze-contingent rendering that renders the gaze central region with high quality while decreasing the fidelity in peripheral region to increase rendering efficiency~\cite{patney2016towards}; 2) gaze-based interaction that uses eye gaze to interact with 3D objects in virtual environments~\cite{sidenmark2019eyehead}; 3) virtual content optimisation that optimises the design of virtual content based on gaze distributions~\cite{alghofaili2019optimizing}; 4) gaze-guided redirected walking that redirects a user's walking path based on their gaze directions~\cite{sun2018towards}; or 5) eye-based activity recognition that recognises user activities from their eye movements~\cite{hu2022ehtask}.

% Motivation for eye-body coordination analysis
Understanding and predicting human eye movements in VR/AR is challenging given the highly variable spatio-temporal dynamics of gaze allocation as well as various top-down influences, e.g. specific tasks that a user has to perform~\cite{hu2021fixationnet, hu2022ehtask} or social norms that a user ideally adheres to during interactions with humans~\cite{canigueral2019being}.
%\andreas{"agent" falls a bit out of the blue here -- I tried to alleviate it a bit by replacing "system" with "agent" above. but you might want to say more explicitly that the applications you mention in detail are for agents interacting with humans in AR/VR.}
Research on gaze behaviour analysis has revealed that eye movements are closely coordinated with head movements and, as such, that head orientation can be used as a proxy to eye gaze \cite{sitzmann2018saliency,hu2019sgaze,hu2020dgaze,hu2021fixationnet}.
Consequently, previous works have mainly focused on exploiting eye-head coordination for predicting eye movements~\cite{hu2019sgaze,hu2020dgaze,hu2021fixationnet} while neglecting coordination between eye and full-body movements.
Furthermore, most existing works have studied human-object interactions~\cite{hu2021fixationnet,hu2022ehtask} and have neglected the more challenging but for VR/AR applications also more practically useful human-human interactions, in which a person's eye gaze is influenced by the body movements of the interaction partner.

%However, we argue that human eye movements may also have strong correlations with full-body movements in everyday activities. For example, to pick up an object, humans usually fixate on the object while stretching out their hands to grab the object.

% Our work
To fill this gap, in this work we are the first to report comprehensive analyses of the correlations between gaze direction and full-body movements using four publicly available datasets collected in the real world (MoGaze~\cite{kratzer2020mogaze}), in VR (ADT~\cite{pan2023aria}) and AR  (GIMO~\cite{zheng2022gimo} and EgoBody~\cite{zhang2022egobody}).
These datasets contain human eye and full-body motion data recorded using a body tracking system and an eye tracker during various daily activities.
%To the best of our knowledge, we are first to study the coordination of eye movements and full-body movements simultaneously.
Our analyses reveal that in human-object interactions, e.g. \textit{pick}, \textit{place}, \textit{touch}, or \textit{hold}, eye gaze exhibits strong correlations with full-body motion and eye movements precede body movements.
In contrast, in human-human interactions, e.g. \textit{chat}, \textit{teach}, or \textit{discuss}, a person’s gaze is more closely correlated with the body orientation towards the interaction partner.
%with the directions pointing from his body to the body of the interaction partner and that
%and body movements are generally faster than eye movements.
Based on these findings we further present \textit{Pose2Gaze}~-- the first method to predict eye movements from human full-body poses during human-object and human-human interactions.
At the core of our method is a novel learning-based eye-body coordination model that based on a convolutional neural network (CNN) and a spatio-temporal graph convolutional neural network (GCN) to extract features from head directions and full-body poses, respectively.
A convolutional neural network is also used to predict human eye gaze from these extracted features.
We compare our method with state-of-the-art gaze prediction methods for three different prediction tasks, i.e. generating target eye gaze from past, present, and future body poses, respectively, and show that our method outperforms these baselines by a large margin in terms of mean angular error on the MoGaze ($24.0\%$ improvement), ADT ($10.1\%$), GIMO ($21.3\%$), and EgoBody ($28.6\%$) datasets.
We also evaluate the effectiveness of our method for a sample downstream task, i.e eye-based activity recognition, and demonstrate that using our gaze prediction method results in significant improvements in recognition accuracy. The full source code and trained models are available at zhiminghu.net/hu24\_pose2gaze.
%\footnote{The full source code and trained models are available at https://zhiminghu.net/hu24\_pose2gaze}

\vspace{1em}
\noindent
The specific contributions of our work are four-fold:
\begin{itemize}

\item We provide comprehensive analyses of eye-body coordination in human-object and human-human interactions in the real world, VR, and AR, and show that eye gaze is strongly correlated with full-body motion in human-object interactions and is closely linked with the directions between two bodies in human-human interactions.

\item We propose \textit{Pose2Gaze} -- a novel eye-body coordination model for gaze prediction that combines a CNN and a spatio-temporal GNN to extract features from head directions and full-body poses.

\item We report extensive experiments on four public datasets for three different prediction tasks and demonstrate significant performance improvements over several state-of-the-art methods.

\item We demonstrate our method's effectiveness in the sample downstream task of eye-based activity recognition.
\end{itemize}

%% file: sections/related_work.tex
\section{Related Work}
%Our work is related to previous research on 1) human movement prediction, 2) eye-body coordination, as well as 3) eye gaze prediction.
\subsection{Human Movement Prediction}
Human movement prediction is an important research topic in the areas of virtual reality and augmented reality.
Some researchers focused on predicting human body movements from speech signals.
For example, Hasegawa et al. used recurrent neural networks (RNNs) to generate natural human full-body poses from perceptual features extracted from the input speech audio\cite{hasegawa2018evaluation} while Kucherenko et al. employed autoencoder neural networks to learn latent representations for human poses and speech signals and then learned the mappings between the two representations to predict human body motions\cite{kucherenko2019analyzing}.
Other researchers used the text transcripts of speech to generate human poses.
Specifically, Yoon et al. proposed to use a RNN-based encoder to extract features from speech text and a RNN-based decoder to generate human poses\cite{yoon2019robots} while Bhattacharya et al. employed an end-to-end Transformer network to predict human movements from the text transcripts of speech\cite{bhattacharya2021text2gestures}.
In addition, there also exist some works that generated human dance motions from input music and achieved good performances\cite{ye2020choreonet,lee2019dancing}.
While previous works typically focused on generating human body movements, they neglected the prediction of human eye gaze, which is significant for human-human\cite{higuch2016can,duarte2018action} and human-computer interactions\cite{duchowski2018gaze,sidenmark2019eyehead,jiao23supreyes}.
In this work, we focus on predicting human eye movements directly from human full-body poses to fill this gap.
%Our approach could be integrated with existing body motion generation methods to obtain realistic eye and full-body movements simultaneously.

\subsection{Eye-body Coordination}

The coordination of human eye and body movements is an important research topic in the areas of cognitive science and human-centred computing.
Many researchers focused on the coordinated movements between the eyes and the head\cite{hu2019sgaze, stahl1999amplitude, fang2015eye}.
Specifically, Stahl analysed the process of gaze shifts and found its amplitude to be proportional to that of head movements\cite{stahl1999amplitude}.
Fang et al. further examined eye-head coordination during visual search in a large visual field and found that eye-head coordination plays an important role in visual cognitive processing\cite{fang2015eye}.
Hu et al. revealed that human eye movements are strongly correlated with head movements in both free-viewing\cite{hu2019sgaze,hu2020dgaze} and task-oriented settings\cite{hu2021fixationnet,hu2022ehtask} in immersive virtual environments while Kothari et al. identified the coordination of eye and head movements in real-world daily activities\cite{kothari2020gaze}.
Recently, some researchers went beyond eye-head coordination to investigate the correlations between eye movements and the movements of different body parts.
For example, Sidenmark et al. focused on the gaze shift process in immersive virtual reality and discovered general eye, head, and torso coordination patterns\cite{sidenmark2019eye}.
Batmaz et al. developed a VR training system based on the coordination of eye and hand movements and compared user performance in three different virtual environments \cite{batmaz2020touch}.
Emery et al. collected a large-scale dataset that contains human eye, hand, and head movements in a virtual environment and identified the coordination of eye, hand, and head motions\cite{emery2021openneeds}.
Randhavane et al. investigated the effectiveness of eye-gait coordination on expressing emotions and further employed gaze and gait features to generate various emotions for virtual agents\cite{randhavane2019eva}.
However, prior works typically focused on the correlations between eye gaze and a particular body part, e.g. head, hand, or torso.
In contrast, we are the first to study the coordinations of eye and full-body movements simultaneously, which paves the way for predicting eye gaze from full-body poses.

\subsection{Eye Gaze Prediction}

Eye gaze prediction is a popular research topic in the areas of computer vision and human-centred computing. % and many gaze prediction models have been proposed over the last decades.
Typical gaze prediction methods can be classified into bottom-up and top-down approaches.
Bottom-up methods predict eye gaze from the low-level image features of the scene content such as intensity, colour, and orientation\cite{itti1998model, cheng2015global}.
For example, Itti et al. used multiscale colour, intensity, and orientation features extracted from the image to predict saliency map (density map of eye gaze distribution)\cite{itti1998model} while Cheng et al. employed the global contrast features of the image to generate saliency map\cite{cheng2015global}.
Top-down approaches take high-level features of the scene such as specific tasks and context information into consideration to predict eye gaze\cite{borji2012probabilistic, koulieris2016gaze}.
For example, Borji et al. employed players' input such as 2D mouse positions and joystick buttons to predict their eye gaze\cite{borji2012probabilistic} while Koulieris et al. used game state variables to predict users' eye gaze positions in video games\cite{koulieris2016gaze}.
In addition to the typical bottom-up and top-down approaches, recently some researchers took the eye-head coordination into consideration and attempted to predict eye gaze from human head movements\cite{sitzmann2018saliency,hu2019sgaze}.
Specifically, Nakashima et al. proposed to use head directions as prior knowledge to improve the accuracy of bottom-up saliency prediction methods through simple multiplication of the predicted saliency map by a Gaussian head direction bias\cite{nakashima2015saliency}.
Sitzmann et al. employed users' head orientations to predict saliency maps for 360-degree images and achieved an accuracy that is on par with the performance of bottom-up saliency predictors\cite{sitzmann2018saliency}.
Hu et al. proposed to use users' head rotation velocities to predict users' eye gaze positions in immersive virtual environments and achieved good performances
% in both free-viewing\cite{hu2019sgaze,hu2020dgaze} and task-oriented situations
\cite{hu2021fixationnet,hu2021eye}.
However, existing works have only explored the effectiveness of head movements on the task of eye gaze prediction and were limited to human-object interactions.
In stark contrast, in this work we %focus on the coordination of eye and full-body movements and
investigate the effectiveness of full-body movements on predicting human eye gaze in both human-object and human-human interaction activities.

%% file: sections/analysis.tex
\section{Analysis of Eye-body Coordination} \label{sec:analysis}

\subsection{Gaze and Pose Data} \label{sec:datasets}
\hl{We studied the eye-body coordination from two perspectives, i.e. the correlation between gaze direction and body orientations and the correlation between gaze direction and body motions (translational movements of the body).}
We conducted a comprehensive analysis based on four public datasets that contain various human-object and human-human interaction activities collected from the real world (MoGaze~\cite{kratzer2020mogaze}), VR environments (ADT~\cite{pan2023aria}), as well as AR scenarios (GIMO~\cite{zheng2022gimo} and EgoBody~\cite{zhang2022egobody}).

\paragraph{MoGaze}
The MoGaze dataset contains 180 minutes of full-body motion capture data collected from seven participants performing everyday \textit{pick} and \textit{place} activities in an indoor environment.
One participant's eye gaze data is not recorded. Therefore, we only used the data from six people to analyse eye-body coordination.

\paragraph{ADT} The ADT dataset collects human eye gaze and/or full-body pose data performing various indoor activities in two virtual environments (an apartment and an office environment).
To analyse eye-body coordination, we used the $34$ sequences that contain both eye gaze and full-body pose data.
Each sequence lasts for around 2 minutes and the activities include \textit{decoration}, \textit{meal}, and \textit{work}.

\paragraph{GIMO} The GIMO dataset records eye gaze and full-body poses from $11$ people performing various daily activities in various indoor scenes.
The whole dataset contains $215$ sequences and each sequence lasts for around 10 seconds.
The activities can be classified into three categories, i.e. \textit{change the state of objects} (\textit{open}, \textit{push}, \textit{transfer}, \textit{throw}, \textit{pick up}, \textit{lift}, \textit{connect}, \textit{screw}, \textit{grab}, \textit{swap objects}), \textit{interact with objects} (\textit{touch}, \textit{hold}, \textit{step on}, \textit{reach to objects}), and \textit{rest} (\textit{sit} or \textit{lay on objects}).
The original dataset does not provide activity labels for each sequence.
To gain a comprehensive analysis, we manually labelled each sequence as one of the three categories.

\paragraph{EgoBody}
The EgoBody dataset collects eye gaze and full-body pose data from $36$ people performing diverse human-human interaction activities (between two humans) in 15 indoor scenes.
The whole dataset contains $125$ sequences and each sequence lasts for around 2 minutes.
The activities include \textit{catch}, \textit{chat}, \textit{dance}, \textit{discuss}, \textit{learn}, \textit{perform}, and \textit{teach}.

\subsection{Correlation between Eye Gaze and Body Orientations}\label{sec:gaze_body_orientation}

% Angular error of prediction results.
\begin{table}[t] 
	\centering
        % set the caption of the table.
	\caption{The cosine similarities between eye gaze direction and the directions of different body joints in the MoGaze, GIMO, and EgoBody datasets. Gaze direction is strongly correlated with body orientations, especially with head direction.} \label{tab:gaze_body_orientation}
        \resizebox{0.5\textwidth}{!}{
	\begin{tabular}{cccccccc}
		\toprule
        & & \textit{base} & \textit{pelvis} & \textit{torso} & \textit{neck} & \textit{head} \\ \hline
  \multirow{2}{*}{\textit{MoGaze}} &\textit{pick} & 0.64 & 0.60 & 0.66 & 0.84 & 0.92\\ 
    &\textit{place} & 0.62 & 0.58 & 0.63 & 0.84 & 0.92\\ \midrule
\multirow{3}{*}{\textit{GIMO}} &\textit{change} & 0.76 & 0.86 & 0.86 & 0.90 & 0.93 \\ 
    &\textit{interact} & 0.72 & 0.82 & 0.83 & 0.87 & 0.93\\ 
    &\textit{rest} & 0.67 & 0.82 & 0.83 & 0.87 & 0.92\\ 
    \midrule
  \multirow{7}{*}{\textit{EgoBody}} & \textit{catch} &0.90 &0.94 &0.94 &0.96 &0.97\\ 
  & \textit{chat} &0.81 &0.85 &0.87 &0.90 &0.94\\ 
  & \textit{dance} &0.82 &0.86 &0.87 &0.93 &0.97\\ 
  & \textit{discuss} &0.88 &0.88 &0.91 &0.93 &0.94\\ 
  & \textit{learn} &0.70 &0.75 &0.77 &0.84 &0.89\\ 
  & \textit{perform} &0.90 &0.92 &0.92 &0.95 &0.97\\ 
  & \textit{teach} &0.84 &0.84 &0.86 &0.89 &0.93\\
        \bottomrule
	\end{tabular}}	
\end{table}

Prior works have mainly focused on the correlation between eye gaze and head directions~\cite{hu2019sgaze, hu2020dgaze}.
To gain a comprehensive understanding of eye-body coordination, we simultaneously analysed the correlation between eye gaze and the orientations of different body joints, e.g. \textit{neck} and \textit{torso}.
The ADT dataset does not provide the orientations of the body joints and therefore we 
only performed analysis on the MoGaze, GIMO, and EgoBody datasets.
Specifically, we employed the forward directions of the body joints to represent body orientations and used the cosine similarity to analyse the correlation between eye gaze and body orientations.
Cosine similarity measures the similarity between two non-zero vectors by calculating the cosine of the angle between the vectors and produces a value in the range from $-1$ to $+1$, where $-1$ indicates perfect negative correlation, $0$ denotes no correlation, and $+1$ represents perfect positive correlation.
\autoref{tab:gaze_body_orientation} summarises the cosine similarities between eye gaze direction and the directions of different body joints in the MoGaze, GIMO, and EgoBody datasets.
We can see that most of the cosine similarities are larger than $0.6$, demonstrating that eye gaze has very high correlations with body orientations in various daily activities.
In particular, we find that in all the activities eye gaze consistently exhibits highest correlation with head direction ($> 0.9$ in most cases), validating that head direction correlates better with eye gaze than other body orientations.

\begin{figure}[t]
    \centering
    \subfloat[MoGaze]{\includegraphics[width=0.49\linewidth]{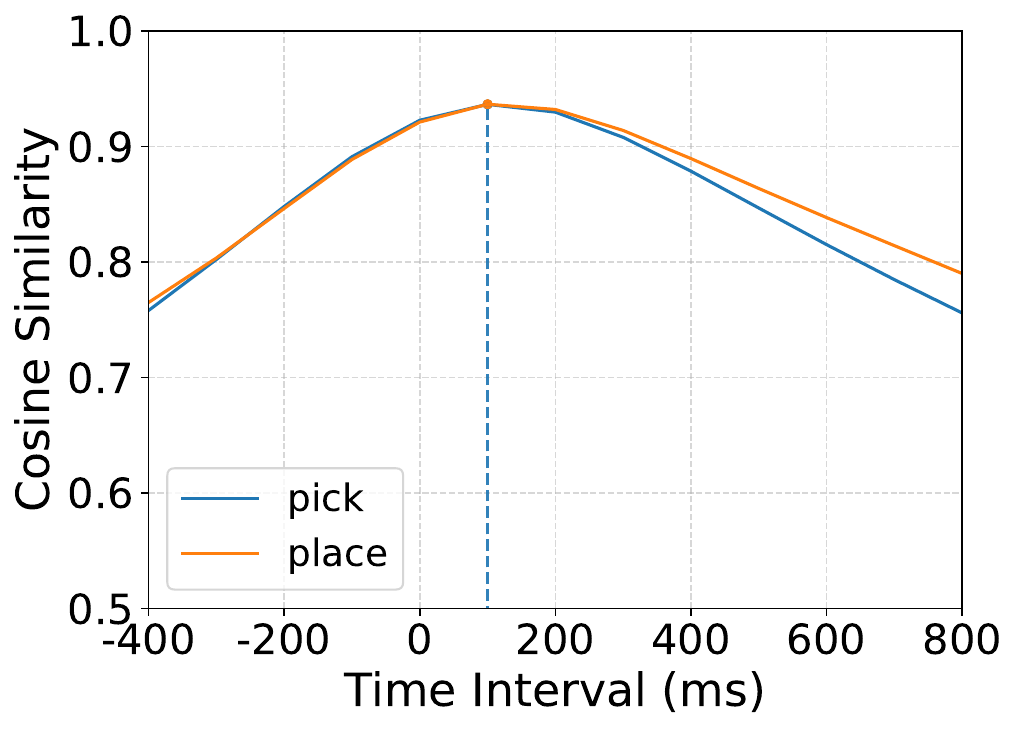}} 
    \subfloat[ADT]{\includegraphics[width=0.49\linewidth]{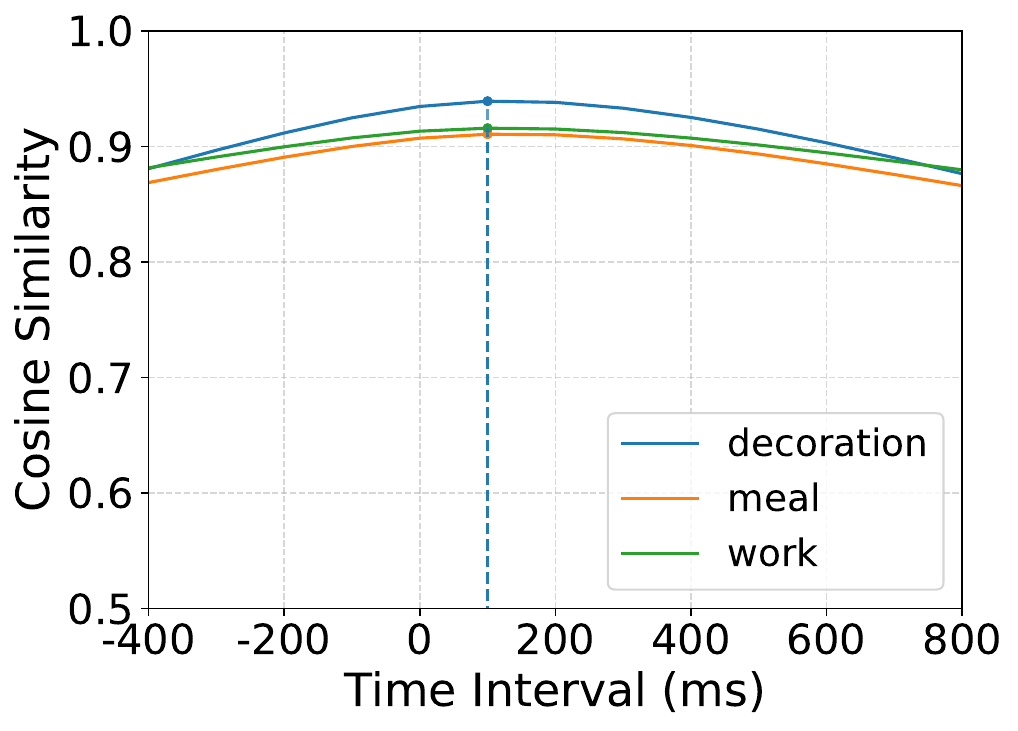}}\newline
    \subfloat[GIMO]{\includegraphics[width=0.49\linewidth]{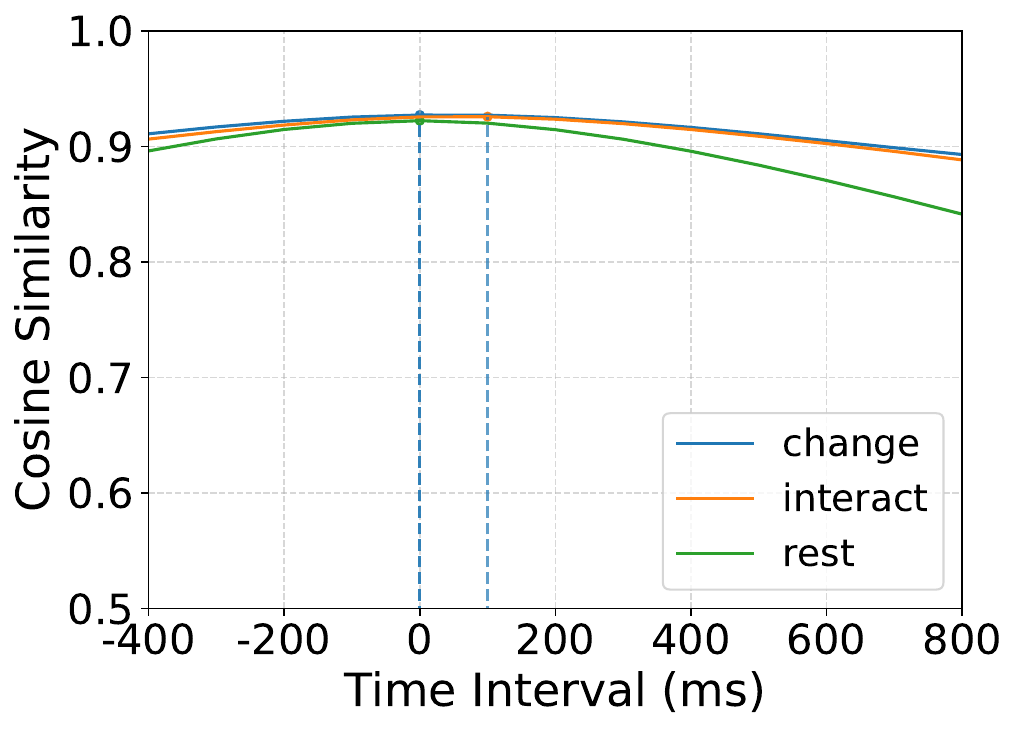}} 
    \subfloat[EgoBody]{\includegraphics[width=0.49\linewidth]{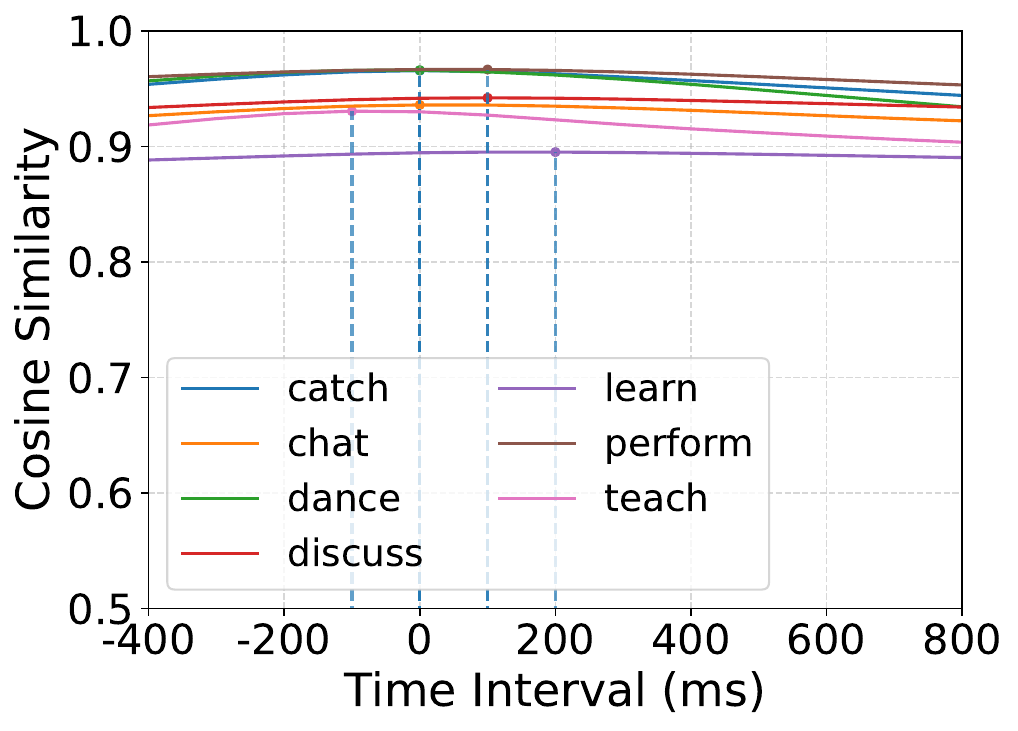}}
    \caption{The cosine similarities between head and gaze directions at different time intervals in the (a) MoGaze (b) ADT (c) GIMO and (d) EgoBody datasets. The highest correlations occur at between $-100$ and $200$ $ms$, suggesting that there is little or no time delay between head and eye movements.}\label{fig:gaze_head_direction}
\end{figure}

To investigate whether there exist time delays between eye gaze and head orientations, we desynchronised the gaze and head signals by adding different time intervals between them (time interval $> 0$ means head data is moved forward in time while time interval $< 0$ means gaze data is moved forward) and further calculated their cosine similarities.
We can see from \autoref{fig:gaze_head_direction} that head direction achieves its highest correlation with eye gaze at the time interval of between $-100$ and $200$ $ms$, indicating that there exists little or no time delay between head and eye movements.
These results further validate that head direction is a suitable proxy for eye gaze in everyday activities.

% figures, [htbp] is used to set the format of our figures.

\subsection{Correlations between Eye Gaze and Body Motions}\label{sec:gaze_body_motion}

\hl{
%We used the term ``body motion'' to indicate the translational movement of the body.
To analyse the correlations between human eye gaze and human full-body motion, we calculated the velocities of different body joints using the difference in position between two consecutive frames and then normalised these velocities to 3D unit vectors to represent the directions of body motions.}
We further calculated the cosine similarities between eye gaze and the directions of body motions to measure their correlations.
\autoref{tab:gaze_body_motion} shows the cosine similarities between eye gaze and full-body motions in the MoGaze, ADT, GIMO, and EgoBody datasets.
%The correlations of the left and right body joints were averaged for simplicity, e.g. \textit{left wrist} and \textit{right wrist} were averaged into \textit{wrist}.
We find that in the human-object interaction activities, e.g. \textit{pick}, \textit{place}, \textit{decoration}, \textit{change}, and \textit{interact}, eye gaze exhibits strong correlations with full-body motions ($>$ 0.3 in most cases).
In the \textit{meal} and \textit{work} activities, the eye-body correlations are relatively smaller, probably because the body motions are less frequent in such activities and are thus less correlated with eye gaze.
In the human-human interaction activities, e.g. \textit{chat}, \textit{discuss}, and \textit{learn}, we find that eye gaze has little or no correlation with body motions ($\leq$ 0.05 in most cases).
This is probably because a person's eye and body motion are strongly influenced by the interaction partner in human-human interactions, thus degrading the eye-body correlations.

% Angular error of prediction results.
\begin{table*}[t] 
	\centering
        \setlength{\tabcolsep}{2pt}
        % set the caption of the table.
	\caption{The cosine similarities between eye gaze and the motions of different body joints in the MoGaze, ADT, GIMO, and EgoBody datasets. Eye gaze has strong correlations with body motions in human-object interaction activities while having little or no correlation in human-human interaction activities. \textit{l\_col: left collar, r\_col: right collar, l\_sho: left shoulder, r\_sho: right shoulder, l\_elb: left elbow, r\_elb: right elbow, l\_wri: left wrist, r\_wri: right wrist, l\_hip: left hip, r\_hip: right hip, l\_kne: left knee, r\_kne: right knee, l\_ank: left ankle, r\_ank: right ankle, l\_toe: left toe, r\_toe: right toe.}} \label{tab:gaze_body_motion}
        \resizebox{1.0\textwidth}{!}{
	\begin{tabular}{cccccccccccccccccccccccc}
		\toprule
        & & \textit{base} & \textit{pelvis} & \textit{torso} & \textit{neck} & \textit{head} & \textit{l\_col} & \textit{r\_col} &\textit{l\_sho} &\textit{r\_sho} &\textit{l\_elb} &\textit{r\_elb} &\textit{l\_wri} &\textit{r\_wri} &\textit{l\_hip} &\textit{r\_hip} & \textit{l\_kne} & \textit{r\_kne} &\textit{l\_ank} &\textit{r\_ank} &\textit{l\_toe} &\textit{r\_toe} &Average\\ \hline
  \multirow{2}{*}{\textit{MoGaze}} &\textit{pick} &0.40 &0.40 &0.41 &0.42 &0.46 &0.42 &0.42 &0.38 &0.41 &0.35 &0.40 &0.34 &0.46 &0.40 &0.40 &0.42 &0.42 &0.31 &0.32 &0.37 &0.37 &0.39\\ 
    &\textit{place} &0.48 &0.49 &0.49 &0.50 &0.54 &0.50 &0.50 &0.47 &0.48 &0.44 &0.47 &0.43 &0.58 &0.49 &0.48 &0.50 &0.50 &0.39 &0.39 &0.45 &0.45 &0.48\\ \midrule
\multirow{3}{*}{\textit{ADT}}
    &\textit{decoration} &0.28 &0.28 &0.26 &0.26 &0.27 &0.26 &0.26 &0.26 &0.24 &0.27 &0.23 &0.31 &0.25 &0.28 &0.28 &0.28 &0.26 &0.15 &0.12 &0.20 &0.14 &0.25\\ 
    &\textit{meal} &0.20 &0.20 &0.20 &0.20 &0.20 &0.20 &0.20 &0.20 &0.19 &0.19 &0.20 &0.22 &0.22 &0.20 &0.20 &0.20 &0.20 &0.09 &0.08 &0.13 &0.10 &0.18\\ 
    &\textit{work} &0.18 &0.19 &0.19 &0.20 &0.22 &0.20 &0.20 &0.20 &0.18 &0.19 &0.17 &0.20 &0.18 &0.18 &0.18 &0.19 &0.18 &0.10 &0.09 &0.14 &0.10 &0.17\\\midrule
  \multirow{3}{*}{\textit{GIMO}} & \textit{change} &0.34 &0.34 &0.35 &0.35 &0.34 &0.35 &0.35 &0.34 &0.34 &0.33 &0.33 &0.29 &0.32 &0.34 &0.34 &0.31 &0.31 &0.19 &0.17 &0.15 &0.10 &0.30\\ 
  & \textit{interact} &0.38 &0.38 &0.37 &0.37 &0.36 &0.37 &0.37 &0.36 &0.36 &0.35 &0.36 &0.32 &0.36 &0.38 &0.38 &0.35 &0.34 &0.21 &0.21 &0.18 &0.15 &0.33\\
  & \textit{rest} &0.36 &0.35 &0.35 &0.34 &0.34 &0.35 &0.35 &0.34 &0.34 &0.32 &0.32 &0.30 &0.32 &0.36 &0.37 &0.33 &0.33 &0.20 &0.18 &0.17 &0.14 &0.31\\ \midrule
  \multirow{7}{*}{\textit{EgoBody}} &\textit{catch} &0.03 &0.02 &0.02 &0.01 &0.02 &0.02 &0.01 &0.03 &0.00 &0.03 &-0.02 &0.04 &0.00 &0.03 &0.02 &0.02 &0.02 &0.02 &0.00 &0.02 &0.01 &0.02\\
    &\textit{chat} &0.01 &0.01 &0.01 &0.02 &0.02 &0.01 &0.01 &0.02 &0.02 &0.01 &0.01 &0.01 &0.02 &0.01 &0.01 &0.00 &0.01 &0.01 &0.01 &0.01 &0.01 &0.01\\
    &\textit{dance} &0.05 &0.05 &0.05 &0.04 &0.04 &0.04 &0.05 &0.04 &0.04 &0.03 &0.04 &0.03 &0.03 &0.05 &0.05 &0.05 &0.04 &0.02 &0.01 &0.02 &0.02 &0.04\\
    &\textit{discuss} &0.02 &0.02 &0.03 &0.03 &0.04 &0.03 &0.03 &0.04 &0.03 &0.02 &0.03 &0.03 &0.03 &0.01 &0.01 &0.00 &0.02 &0.00 &0.00 &0.01 &0.01 &0.02\\ 
    &\textit{learn} &0.00 &-0.01 &-0.01 &-0.01 &-0.01 &-0.01 &0.00 &-0.01 &0.00 &-0.01 &-0.01 &-0.01 &0.00 &0.00 &0.00 &-0.01 &-0.01 &0.00 &0.01 &0.01 &0.01 &0.00\\ 
    &\textit{perform} &0.04 &0.04 &0.02 &0.02 &0.01 &0.01 &0.03 &-0.01 &0.01 &-0.03 &0.01 &0.01 &0.02 &0.04 &0.05 &0.03 &0.01 &0.01 &0.03 &0.02 &0.03 &0.02\\
    &\textit{teach} &0.00 &0.00 &0.01 &0.01 &0.02 &0.01 &0.01 &0.00 &0.02 &0.00 &0.02 &0.01 &0.02 &0.00 &0.01 &0.01 &0.02 &0.02 &0.02 &0.02 &0.02 &0.01\\
        \bottomrule
	\end{tabular}}	
\end{table*}

\begin{figure}[t]
    \centering
    \subfloat[MoGaze]{\includegraphics[width=0.49\linewidth]{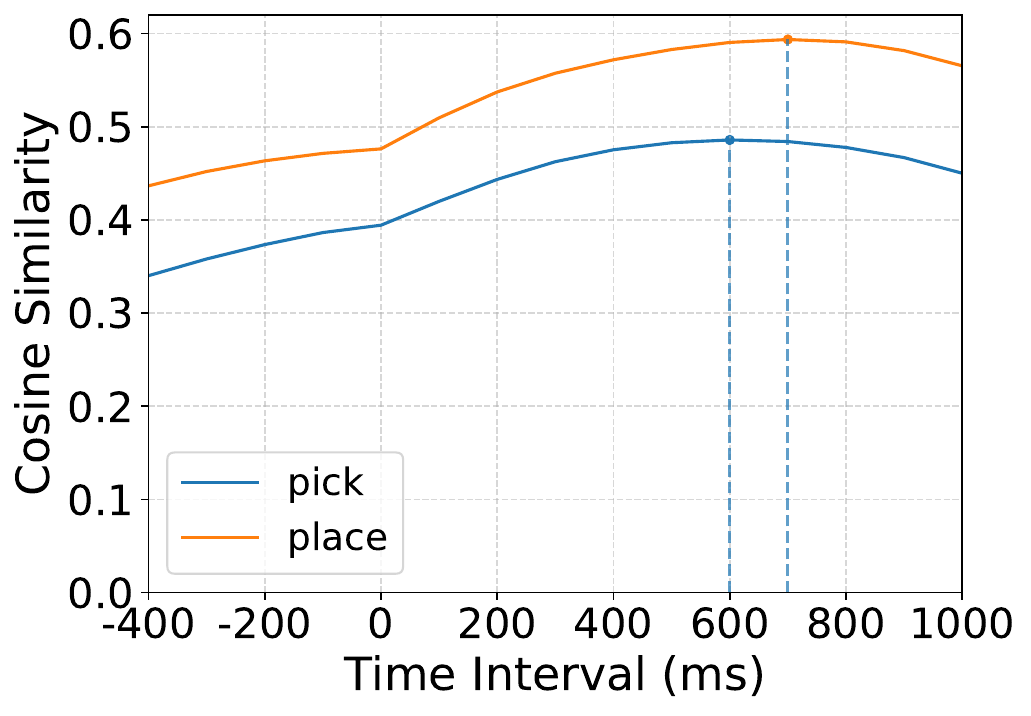}} 
    \subfloat[ADT]{\includegraphics[width=0.49\linewidth]{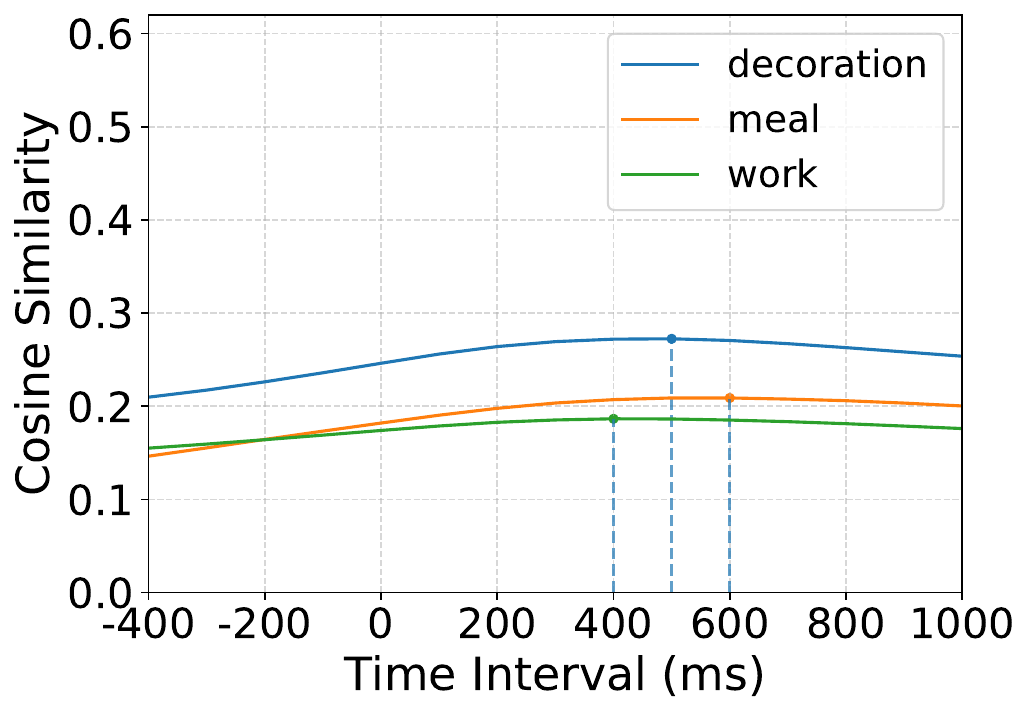}}\newline
    \subfloat[GIMO]{\includegraphics[width=0.49\linewidth]{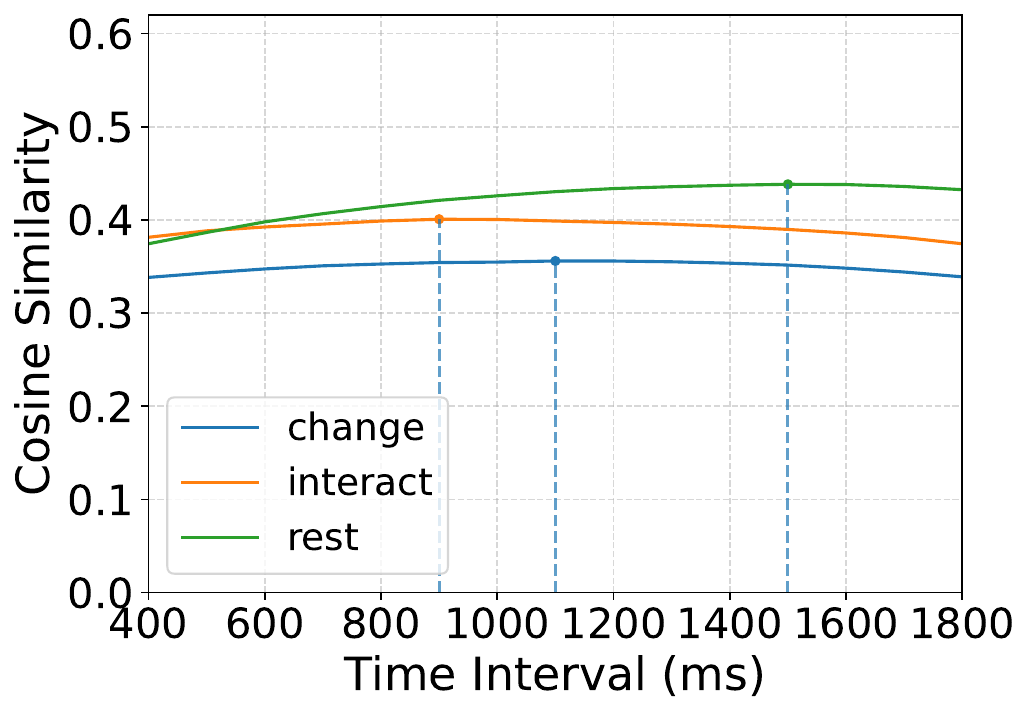}}
    \subfloat[EgoBody]{\includegraphics[width=0.49\linewidth]{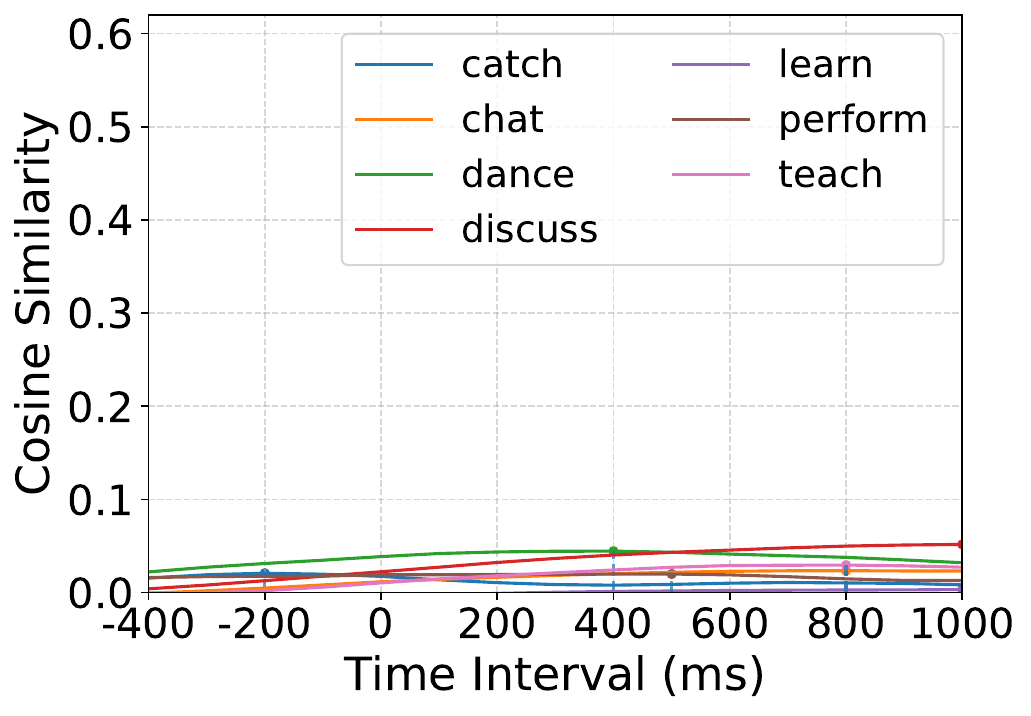}}
    \caption{The cosine similarities between eye gaze and body motions at different time intervals in the (a) MoGaze (b) ADT (c) GIMO and (d) EgoBody datasets. The highest correlations occur at between $400$ and $1500$ $ms$, indicating that eye movements precede body motions.}\label{fig:gaze_body_motion}
\end{figure}

To investigate any potential time delays between body motions and eye movements, we added different time intervals between body motions and gaze directions (time interval $> 0$ means body motion data is moved forward in time while time interval $< 0$ means gaze data is moved forward) and further calculated their correlations.
Specifically, we first calculated the cosine similarities between eye gaze and the motions of different body joints and then averaged the cosine similarities across all the body joints.
We can see from \autoref{fig:gaze_body_motion} that in human-object interaction activities the highest correlations occur at between $400$ and $1500$ $ms$, suggesting that there exists a noticeable time delay between body motions and eye movements, i.e. eye movements precede body motions.
In the human-human interaction activities, we find that the eye-body correlations are consistently small ($<$ 0.1 in all the cases) at different time intervals, validating that there is little or no correlation between eye gaze and body motion in human-human interactions.

% figures, [htbp] is used to set the format of our figures.

\subsection{Eye-body Coordination in Human-human Interactions}\label{sec:gaze_two_body_motion}

\begin{figure}[t]
    \centering
    \subfloat[]{\includegraphics[height=0.4\linewidth]{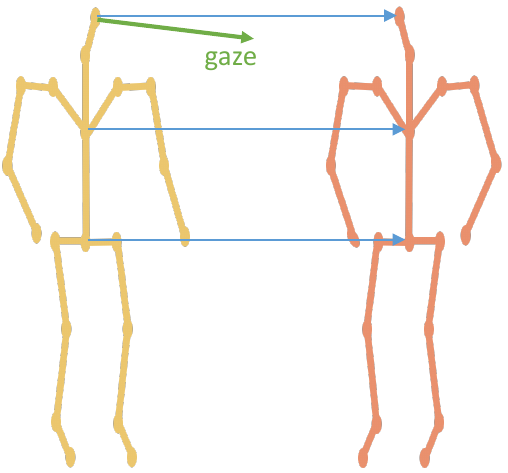}}
    \subfloat[]{\includegraphics[height=0.4\linewidth]{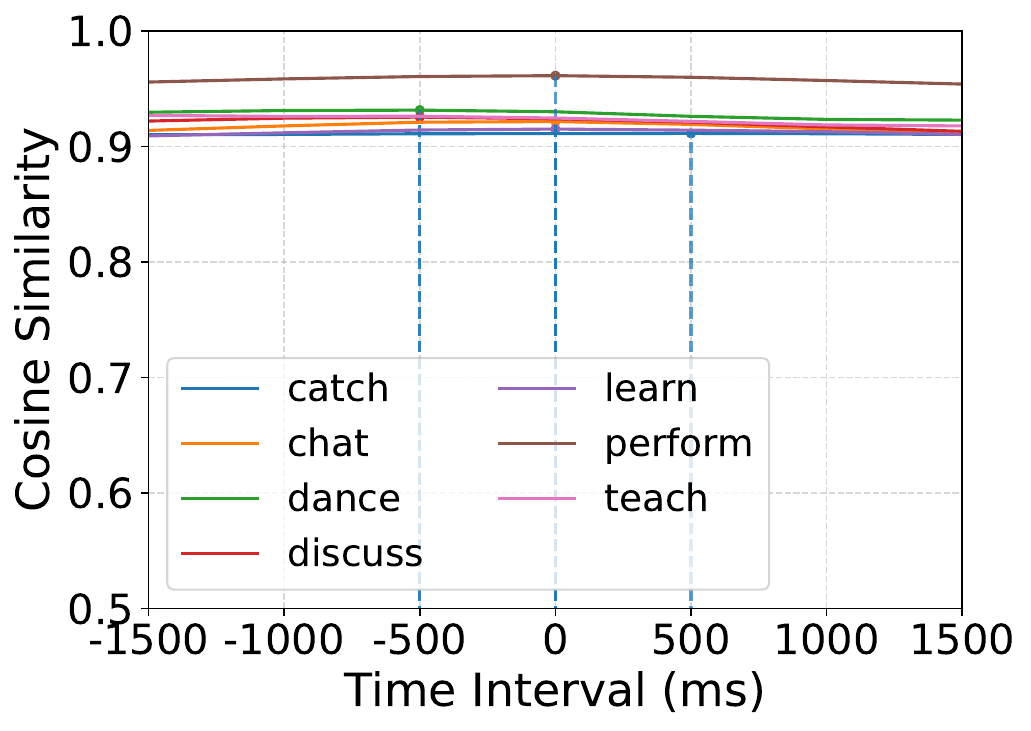}}
    \caption{(a) Eye gaze and the directions pointing from a person's body to the body of the interaction partner and (b) the cosine similarities between eye gaze and the directions between two bodies at different time intervals.
    \hl{The highest correlations either occur at $0$ $ms$ or are very close to the correlation values at $0$ $ms$, suggesting that there is little or no time delay between body motions and eye gaze.}}\label{fig:gaze_two_body_motion}
\end{figure}

To analyse the influence of the interaction partner on a person's eye gaze during human-human interactions, we calculated the directions pointing from a person's body joints to the body joints of the interaction partner (see \autoref{fig:gaze_two_body_motion}a).
To this end we used the difference in position between the person and the interaction partner and normalised these directions to 3D unit vectors.
We further calculated the cosine similarities between eye gaze and the directions between two bodies to measure their correlation.
We can see in \autoref{tab:gaze_two_body_motion} that eye gaze is highly correlated with the directions between both bodies ($>0.9$ in most cases), suggesting that a person's eye gaze is significantly influenced by the interaction partner in human-human interactions.
\autoref{fig:gaze_two_body_motion}b further shows the cosine similarities between eye gaze and the directions between two bodies at different time intervals.
We can see that the highest correlations either occur at $0$ $ms$ or are very close to the correlation values at $0$ $ms$, meaning that there is little or no time delay between body motions and eye movements in human-human interactions.

% Angular error of prediction results.
\begin{table*}[t] 
	\centering
        \setlength{\tabcolsep}{2pt}
        % set the caption of the table.
	\caption{The cosine similarities between eye gaze and the directions pointing from a person's body to the interaction partner in human-human interaction activities. Eye gaze is highly correlated with the directions between two bodies. \textit{l\_col: left collar, r\_col: right collar, l\_sho: left shoulder, r\_sho: right shoulder, l\_elb: left elbow, r\_elb: right elbow, l\_wri: left wrist, r\_wri: right wrist, l\_hip: left hip, r\_hip: right hip, l\_kne: left knee, r\_kne: right knee, l\_ank: left ankle, r\_ank: right ankle, l\_toe: left toe, r\_toe: right toe.}} \label{tab:gaze_two_body_motion}
        \resizebox{1.0\textwidth}{!}{
	\begin{tabular}{cccccccccccccccccccccccc}
		\toprule
        & & \textit{base} & \textit{pelvis} & \textit{torso} & \textit{neck} & \textit{head} & \textit{l\_col} & \textit{r\_col} &\textit{l\_sho} &\textit{r\_sho} &\textit{l\_elb} &\textit{r\_elb} &\textit{l\_wri} &\textit{r\_wri} &\textit{l\_hip} &\textit{r\_hip} & \textit{l\_kne} & \textit{r\_kne} &\textit{l\_ank} &\textit{r\_ank} &\textit{l\_toe} &\textit{r\_toe} &Average\\ \hline
\multirow{7}{*}{\textit{EgoBody}} &\textit{catch} &0.92 &0.92 &0.92 &0.92 &0.91 &0.92 &0.92 &0.91 &0.91 &0.90 &0.90 &0.89 &0.89 &0.92 &0.92 &0.91 &0.91 &0.92 &0.91 &0.91 &0.90 &0.91\\
    &\textit{chat} &0.94 &0.94 &0.94 &0.94 &0.94 &0.94 &0.94 &0.92 &0.93 &0.91 &0.91 &0.89 &0.89 &0.93 &0.93 &0.91 &0.92 &0.91 &0.91 &0.89 &0.89 &0.92\\
    &\textit{dance} &0.95 &0.95 &0.95 &0.95 &0.95 &0.95 &0.95 &0.94 &0.94 &0.91 &0.92 &0.87 &0.90 &0.94 &0.95 &0.92 &0.93 &0.91 &0.93 &0.89 &0.92 &0.93\\
    &\textit{discuss} &0.93 &0.93 &0.93 &0.93 &0.94 &0.93 &0.93 &0.93 &0.92 &0.92 &0.90 &0.91 &0.88 &0.93 &0.93 &0.92 &0.92 &0.92 &0.92 &0.91 &0.91 &0.92\\ 
    &\textit{learn} &0.93 &0.92 &0.92 &0.92 &0.92 &0.92 &0.92 &0.92 &0.92 &0.90 &0.91 &0.87 &0.91 &0.92 &0.92 &0.91 &0.92 &0.91 &0.92 &0.89 &0.91 &0.91\\ 
    &\textit{perform} &0.97 &0.97 &0.97 &0.97 &0.97 &0.97 &0.97 &0.97 &0.96 &0.96 &0.95 &0.95 &0.94 &0.97 &0.97 &0.96 &0.95 &0.96 &0.95 &0.96 &0.94 &0.96\\ 
    &\textit{teach} &0.93 &0.93 &0.93 &0.93 &0.93 &0.93 &0.93 &0.92 &0.93 &0.91 &0.92 &0.90 &0.91 &0.93 &0.93 &0.92 &0.93 &0.92 &0.92 &0.91 &0.92 &0.92\\
        \bottomrule
	\end{tabular}}	
\end{table*}

% figures, [htbp] is used to set the format of our figures.

\subsection{Summary}

Through comprehensive analyses we found that in various daily activities, eye gaze is closely correlated with body orientations, especially with head direction, and there is little or no time delay between head and eye movements.
In human-object interactions, we find that eye gaze is strongly correlated with human full-body motions and eye movements precede body motions.
In human-human interactions, we reveal that eye gaze has little or no correlation with body motions while having high correlations with the directions pointing from a person to the interaction partner and that there is little or no time delay between body motions and eye movements.
These results suggest that head directions and full-body poses contain rich information about eye gaze and thus can be used to predict eye movements.

%% file: sections/method.tex
\section{Pose2Gaze: Eye-body Coordination Model}

\subsection{Model Design}
\begin{figure*}[t]
    \centering
    \includegraphics[width=0.9\textwidth]{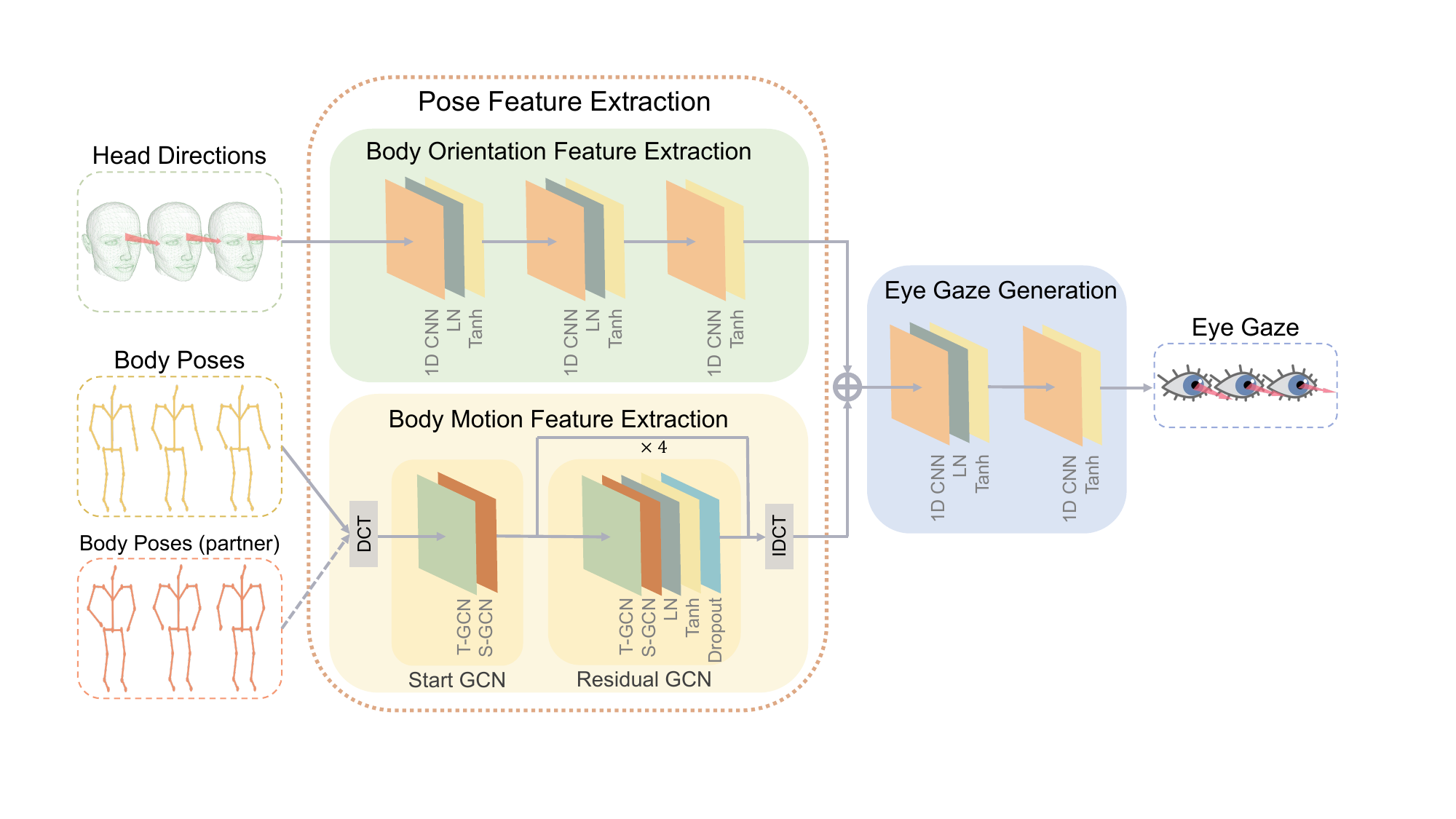}
    \caption{Architecture of the proposed \textit{Pose2Gaze} model.
    \textit{Pose2Gaze} first uses a 1D convolutional neural network to extract body orientation features from head directions, then applies a spatio-temporal graph convolutional neural network to extract the body motion features from human full-body poses, and finally employs a 1D convolutional neural network to generate human eye gaze from the extracted body orientation and motion features.}\label{fig:method}
\end{figure*}

\hl{\paragraph{Design of Pose2Gaze}
The analyses in \autoref{sec:analysis} revealed that eye gaze is closely correlated with body orientations as well as full-body motions.
Based on these insights, we propose an eye-body coordination model for gaze prediction that consists of three main components: a body orientation feature extraction module that extracts orientation features, a body motion feature extraction module that extracts motion features, and an eye gaze generation module that generates gaze directions from the extracted body orientation and motion features (see \autoref{fig:method} for an overview of our method).
In the body orientation feature extraction module, we use head direction as input since it has higher correlation with eye gaze than other body orientations (see \autoref{tab:gaze_body_orientation}).
In the body motion feature extraction module, we use human full-body poses as input in human-object interaction activities since eye gaze is strongly correlated with full-body motions in this situation (see \autoref{tab:gaze_body_motion}) and employ the full-body poses of both the person and their interaction partner as input in human-human interactions considering that eye gaze has high correlations with the directions pointing from a person to the interaction partner in this setting (see \autoref{tab:gaze_two_body_motion}).}

\paragraph{Problem Formulation} We define pose-based eye gaze prediction as the task of generating a sequence of eye gaze directions $G_{t+1:t+T} = \{g_{t+1}, g_{t+2}, ..., g_{t+T}\}\in R^{3\times T}$, where $g$ is a 3D unit vector and $T$ is the length of the target eye gaze sequence, from human body orientations and motions.
We use a sequence of head directions $H_{t+1+\Delta t_h:t+T+\Delta t_h} = \{h_{t+1+\Delta t_h}, h_{t+2+\Delta t_h}, ..., h_{t+T+\Delta t_h}\}\in R^{3\times T}$ to represent body orientations, where $h$ is a 3D unit vector and $\Delta t_h$ is the time interval between the input head directions and target eye gaze.
We employ a sequence of the 3D positions of all human joints $P_{t+1+\Delta t_p:t+T+\Delta t_p} = \{p_{t+1+\Delta t_p}, p_{t+2+\Delta t_p}, ..., p_{t+T+\Delta t_p}\}\in R^{3\times N\times T}$ to represent body motions, where $N$ is the number of human joints and $\Delta t_p$ is the time interval between the input body poses and target eye gaze.
%In human-human interactions, we also use the body poses of the interaction partner as input by simply merging the pose data of the two bodies ($P\in R^{3\times 2N\times T}$, $N$ joints from the person and $N$ joints from the interaction partner).
%By setting different time intervals between the input body movements and the target gaze directions, our model can be trained to generate target eye gaze from past, present, or future body poses, respectively (see \autoref{sec:experimental_settings}).

\subsection{Body Orientation Feature Extraction} \label{sec:body_orientation_feature}

\hl{In light of the good performance of 1D convolutional neural network for processing head movement data~\cite{hu2020dgaze,hu2021fixationnet, hu2022ehtask}}, we employed three 1D CNN layers to extract body orientation features from the sequence of head directions.
Specifically, we used two 1D CNN layers, each with $32$ channels and a kernel size of three, to process the head direction sequence $H\in R^{3\times T}$.
Each CNN layer was followed by a layer normalisation (LN) and a Tanh activation function.
After the two CNN layers, we used a 1D CNN layer with $32$ channels and a kernel size of three, and a Tanh activation function to obtain the body orientation features $f_{ori}\in R^{32\times T}$.

\subsection{Body Motion Feature Extraction}\label{sec:body_motion_feature}
%\paragraph{Pose Data Representation}
%To extract motion features from human full-body poses, we represented human pose data $P\in R^{3\times N \times T}$ as a graph because this representation is beneficial for learning the relationships between human joints~\cite{ma2022progressively,mao2020history}.
%More specifically, we modelled the pose data as a spatial graph and a temporal graph respectively: the spatial graph views each joint data as a node and contains $N$ nodes in total while the temporal graph takes each time step as a node and covers $T$ nodes.
%\andreas{so you have one graph that represents the full body (all joints) -- but how many temporal graphs do you have then? if the temporal graph nodes encode time then one graph for each joint?!}\zhiming{One spatial and one temporal respectively.}
%The spatial and temporal graphs are both fully-connected, with their adjacency matrices measuring the weights between each pair of nodes.

Given the effectiveness of discrete cosine transform (DCT) for extracting temporal features from human pose data~\cite{ma2022progressively,mao2020history}, we first employed DCT to encode human pose $P\in R^{3\times N \times T}$ in the temporal domain using DCT matrix $M_{dct} \in R^{T\times T}$:
\begin{equation} 
\begin{aligned}
    P_{dct} = P M_{dct}, 
\end{aligned}
\end{equation}
where $P_{dct}\in R^{3\times N \times T}$ is the human pose after DCT transform.
%Considering the good performance of graph convolutional neural networks for processing human pose data~\cite{ma2022progressively}
\hl{Considering that graph convolutional network outperforms other architectures like Transformer or RNN for processing human pose data~\cite{ma2022progressively}}, we propose two GCN blocks, i.e. a start GCN block and a residual GCN block, to extract motion features from the transformed pose data.

\paragraph{Start GCN Block}
The start GCN block first applies a temporal GCN (T-GCN) to extract the temporal features from the transformed pose data $P_{dct}$.
The temporal GCN views the pose data as a fully-connected graph that contains $T$ nodes corresponding to pose data at $T$ time steps.
It learns the weighted adjacency matrix $A^T \in R^{T\times T}$ of the fully-connected temporal graph and performs temporal convolution using
\begin{equation} 
\begin{aligned}
    f_{temp} = P_{dct} A^T,
\end{aligned}
\end{equation}
where $f_{temp}\in R^{3\times N\times T}$ is the extracted temporal features.
$f_{temp}$ was then permuted to $f_{temp}\in R^{T\times N\times 3}$.
A weight matrix $W^{start}\in R^{3\times16}$ was applied to convert the input node features ($3$ dimensions) to latent features ($16$ dimensions):
\begin{equation} 
\begin{aligned}
    f_{lat} = f_{temp} W^{start},
\end{aligned}
\end{equation}
where $f_{lat}\in R^{T\times N\times 16}$ is the latent features.
After the weight matrix, a spatial GCN (S-GCN) was applied to extract the spatial features.
The spatial GCN views the latent features $f_{lat}$ as a fully-connected graph that contains $N$ nodes corresponding to $N$ human joints.
S-GCN learns the weighted adjacency matrix $A^S \in R^{N\times N}$ of the fully-connected spatial graph and performs spatial convolution using 
\begin{equation} 
\begin{aligned}
    f_{spa} = A^S f_{lat},
\end{aligned}
\end{equation}
where $f_{spa}\in R^{T\times N\times 16}$ is the extracted spatial features.
$f_{spa}$ was further permuted to $f_{spa}\in R^{16\times N\times T}$.
The output of the spatial GCN is copied along the temporal dimension ($R^{16\times N\times T}\rightarrow R^{16\times N\times 2T}$) to enhance the features~\cite{ma2022progressively} and is then used as input to the residual GCN block.

\paragraph{Residual GCN Block}
The residual GCN block contains $4$ GCN components with each component consisting of a temporal GCN that learns the temporal adjacency matrix $A^T_i \in R^{2T\times 2T}$, a weight matrix $W^{res}_i\in R^{16\times16}$ that extracts the latent features, a spatial GCN that learns the spatial adjacency matrix $A^S_i \in R^{N\times N}$, a layer normalisation, a Tanh activation function, and a dropout layer with dropout rate $0.3$ to avoid overfitting.
We 
%set the number of GCN components $m$ to $4$ and
added a residual connection for each GCN component to improve the network flow.
We further cut the output of the residual GCN block in half in the temporal dimension to reduce the feature dimensions.

\hl{The output of the residual GCN block was converted back to the original representation space using an inverse discrete cosine transform (IDCT)\cite{ma2022progressively}. A Tanh activation function was applied after the IDCT to obtain the spatio-temporal human body motion features $f_{mot}\in R^{16\times N\times T}$.}

\subsection{Eye Gaze Generation}

To generate eye gaze from the extracted body orientation and motion features, we first aggregated the body motion features $f_{mot}$ along the spatial dimension, i.e. concatenated the features of different body joints into a single motion feature ($R^{16\times N\times T}\rightarrow R^{16N\times T}$).
We then fused the extracted orientation and motion features by concatenating them along the spatial dimension and obtained $f\in R^{(16N+32)\times T}$.
We finally used a 1D convolutional neural network to generate eye gaze from the fused features.
Specifically, we used two CNN layers, each with a kernel size of three, to process the fused features.
The first CNN layer has $64$ channels and is followed by a layer normalisation and a Tanh activation function while the second CNN layer uses three channels and a Tanh activation function to generate the eye gaze.
The generated eye gaze directions were finally normalised to unit vectors: $\hat{G}_{t+1:t+T} = \{\hat{g}_{t+1}, \hat{g}_{t+2}, ..., \hat{g}_{t+T}\} \in R^{3\times T}$.

\subsection{Loss Function}

To train our model, we used the mean angular error between the generated eye gaze directions $\hat{G}_{t+1:t+T}$ and the ground truth gaze directions $G_{t+1:t+T}$ as our loss function:
\begin{equation} 
\begin{aligned}
    \ell = \frac{1}{T}\sum_{j=t+1}^{t+T}\arccos(\hat{g_j} \cdot g_j).
\end{aligned}\label{eq:loss}
\end{equation}

%% file: sections/experiments.tex
\section{Experiments and Results} \label{sec:experiments}

We conducted extensive experiments to evaluate the performance of our gaze prediction model.
Specifically, we compared our model with state-of-the-art gaze prediction methods that estimate eye gaze from head movements on the MoGaze, ADT, GIMO, and EgoBody datasets.
We tested these methods under three different generation settings, i.e. generating target eye gaze from past, present, and future body poses, respectively.
We also performed an ablation study to evaluate the effectiveness of each component used in our model.

\subsection{Experimental Settings} \label{sec:experimental_settings}

\paragraph{Datasets}
We evaluated our method on the MoGaze, ADT, GIMO, and EgoBody datasets (see \autoref{sec:datasets} for the details of these datasets).
On the MoGaze dataset, we performed a leave-one-person-out cross-validation to evaluate our model's generalisation capability for different users: We trained on the data from five people and tested on the remaining one, repeated this procedure six times by testing for a different target person, and calculated the average performance across all six iterations.
On the ADT dataset, we randomly selected $24$ sequences for training and used the remaining $10$ sequences for testing.
On the GIMO dataset, we used the default training and test sets provided by the authors~\cite{zheng2022gimo}, i.e. we used data from $12$ scenes for training ($178$ sequences) and data from $14$ scenes ($12$ known scenes and two new scenes) for testing ($37$ sequences).
On the EgoBody dataset, we used $82$ sequences for training and $43$ sequences for testing following the default training and test sets provided in the original paper~\cite{zhang2022egobody}.

\paragraph{Evaluation Metric}
As is commonly used in gaze prediction~\cite{hu2020dgaze, hu2021fixationnet}, we employed the mean angular error between the generated and ground truth gaze directions (see \autoref{eq:loss}) as the metric to evaluate model performance.

\paragraph{Baselines}
We compared our method with the following state-of-the-art gaze prediction methods that generate eye gaze from human head movements:
\begin{itemize}[noitemsep,leftmargin=*]    
    \item \textit{Head Direction}: \textit{Head Direction} has been frequently used as a proxy for eye gaze due to the strong link between eye and head movements~\cite{sitzmann2018saliency,hu2020dgaze,nakashima2015saliency}.    
    \item \textit{DGaze}~\cite{hu2020dgaze}: \textit{DGaze} predicts eye gaze from the sequence of head movements using a 1D convolutional neural network.
    
    \item \textit{FixationNet}~\cite{hu2021fixationnet}: \textit{FixationNet} extracts features from head movement sequence using a 1D convolutional neural network and combines the features with prior knowledge of gaze distribution to generate eye gaze directions.
\end{itemize}

\paragraph{Implementation Details}
We trained the baseline methods from scratch using their default parameters.
To train our method, we used the Adam optimiser with an initial learning rate of $0.005$ that we decayed by $0.95$ every epoch.
We used a batch size of $32$ to train our method for a total of $50$ epochs.
We implemented our method using the PyTorch framework.

\paragraph{Generation Settings}
We set our model to generate $15$ frames (corresponding to $500$ ms) of eye gaze directions $G_{t+1:t+15} = \{g_{t+1}, g_{t+2}, ..., g_{t+15}\}$ following the settings used in prior work~\cite{hu2021fixationnet, hu2020dgaze}.
We evaluated our method for three different generation tasks:  Generating target eye gaze from past, present, and future body poses, respectively:
\begin{itemize}[noitemsep,leftmargin=*]
    \item \textit{Generating Gaze from Past Poses}:
    We used the body poses and head directions in the past $15$ frames $P_{t-14:t} = \{p_{t-14}, p_{t-13}, ..., p_{t}\}$ and $H_{t-14:t} = \{h_{t-14}, h_{t-13}, ..., h_{t}\}$ as input to generate the target eye gaze.
    This setting is equivalent to predicting human eye gaze in the future (gaze forecasting)~\cite{hu2021fixationnet, hu2020dgaze}, which is important for a variety of applications including visual attention enhancement~\cite{el2009dynamic}, dynamic event triggering~\cite{hadnett2019effect}, as well as human-human and human-computer interaction \cite{steil2018forecasting,mueller2020anticipating}.
    
    \item \textit{Generating Gaze from Present Poses}: 
    We used the body poses and head directions at the present time $P_{t+1:t+15} = \{p_{t+1}, p_{t+2}, ..., p_{t+15}\}$ and $H_{t+1:t+15} = \{h_{t+1}, h_{t+2}, ..., h_{t+15}\}$ as input to generate the corresponding gaze directions.    
    This setting corresponds to estimating human eye gaze in real time~\cite{hu2019sgaze, koulieris2016gaze}, which is key for a number of applications including gaze-contingent rendering~\cite{patney2016towards}, gaze-based interaction~\cite{majaranta14_apc}, and gaze-guided redirected walking~\cite{sun2018towards}.
    
    \item \textit{Generating Gaze from Future Poses}: \autoref{sec:gaze_body_motion} reveals that in human-object interactions eye gaze has highest correlations with body motions in the near future and this implies that using future body poses may improve the performance of gaze generation.
    Therefore, we used the body poses in the future $15$ frames $P_{t+16:t+30} = \{p_{t+16}, p_{t+17}, ..., p_{t+30}\}$ and the head directions at the present time $H_{t+1:t+15} = \{h_{t+1}, h_{t+2}, ..., h_{t+15}\}$ as input to generate the target eye gaze.
    We used real-time head directions because there exists little or no time delay between head and eye movements (\autoref{sec:gaze_body_orientation}).
    This setting can be seen as an offline processing way of generating eye movements and is particularly important for the applications that rely on offline analysis of gaze data such as virtual content design and optimisation~\cite{sitzmann2018saliency, alghofaili2019optimizing}.
\end{itemize}

\subsection{Gaze Generation Results} \label{sec:gaze_generation_results}
\paragraph{Generating Gaze from Past Poses}
\autoref{tab:results}\textit{-past} summarises the performances of different methods for generating eye gaze from past body poses on the MoGaze, ADT, GIMO, and EgoBody datasets.
We can see that our method outperforms the state-of-the-art methods in terms of both individual actions as well as all the actions in a dataset, achieving an average improvement of $22.0\%$ ($13.1^\circ$ \textit{vs.} $16.8^\circ$) on MoGaze, $8.0\%$ ($11.5^\circ$ \textit{vs.} $12.5^\circ$) on ADT, $11.1\%$ ($18.4^\circ$ \textit{vs.} $20.7^\circ$) on GIMO, and $28.6\%$ ($13.2^\circ$ \textit{vs.} $18.5^\circ$) on EgoBody.
We further performed a paired Wilcoxon signed-rank test to compare the mean angular error of our method and that of the state-of-the-art methods and validated that the differences between our method and the state of the art are statistically significant ($p<0.01$) on all the four datasets.
In addition, we notice that our method achieves the highest improvement ($28.6\%$) on the EgoBody dataset, demonstrating its effectiveness on the challenging human-human interaction setting.
We also find that our method's improvement on the ADT dataset ($8.0\%$) is relatively smaller compared with the improvements on other datasets.
This is probably because the body motions in the activities of ADT are less frequent and less correlated with eye gaze  (see \autoref{sec:gaze_body_motion}), thus degrading the performance of our method.
The above results demonstrate that our method has a strong capability of predicting eye gaze from past body poses in both human-object and human-human interaction activities.

% Angular error of prediction results.
\begin{table*}[t]
	\centering
	% set the caption of the table.
	\caption{Mean angular errors of different methods for generating eye gaze from past, present, and future body poses on the MoGaze, ADT, GIMO, and EgoBody datasets. Best results are in bold while the second best are underlined.} \label{tab:results} 
        \resizebox{\textwidth}{!}{
	\begin{tabular}{ccccccccccccccccccccc}
		\toprule
    &&\multicolumn{3}{c}{\textbf{MoGaze}} &\multicolumn{4}{c}{\textbf{ADT}} &\multicolumn{4}{c}{\textbf{GIMO}} &\multicolumn{8}{c}{\textbf{EgoBody}}\\ \cmidrule(lr){3-5} \cmidrule(lr){6-9} \cmidrule(lr){10-13} \cmidrule(lr){14-21}
    &&\textit{pick} &\textit{place} &All &\textit{decoration} &\textit{meal} &\textit{work} &All &\textit{change} &\textit{interact} &\textit{rest} &All &\textit{catch} &\textit{chat} &\textit{dance} &\textit{discuss} &\textit{learn} &\textit{perform} &\textit{teach} &All\\ \hline
  \multirow{4}{*}{\textit{past}}&\textit{Head Direction} & $37.8^\circ$ & $34.9^\circ$ & $36.4^\circ$ & $26.5^\circ$ & $30.6^\circ$ & $27.1^\circ$ & $28.0^\circ$ & $23.5^\circ$ & $23.7^\circ$ & $22.9^\circ$ & $23.4^\circ$ &$\underline{14.6}^\circ$ & $18.1^\circ$ & $25.0^\circ$ & $18.0^\circ$ & $17.6^\circ$ & $\underline{16.8}^\circ$ & $24.5^\circ$ & $19.2^\circ$ 
  \\ %\hline
  &\textit{DGaze}\cite{hu2020dgaze} & $18.3^\circ$ & $15.3^\circ$ & $16.9^\circ$ & $\underline{13.6}^\circ$ & $\underline{13.2}^\circ$ & $\underline{11.1}^\circ$ & $\underline{12.5}^\circ$ & $23.1^\circ$ & $\underline{20.4}^\circ$ & $\underline{18.9}^\circ$ & $20.9^\circ$ & $17.1^\circ$ & $17.9^\circ$ & $27.1^\circ$ & $19.6^\circ$ & $17.3^\circ$ & $21.0^\circ$ & $24.6^\circ$ & $19.5^\circ$
  \\ %\hline
  &\textit{FixationNet}\cite{hu2021fixationnet} & $\underline{18.2}^\circ$ & $\underline{15.2}^\circ$ & $\underline{16.8}^\circ$ & $14.8^\circ$ & $14.3^\circ$ & $12.0^\circ$ & $13.5^\circ$ & $\underline{22.2}^\circ$ & $\textbf{20.0}^\circ$ & $19.7^\circ$ & $\underline{20.7}^\circ$ & $15.4^\circ$ & $\underline{17.3}^\circ$ & $\underline{23.7}^\circ$ & $\underline{17.6}^\circ$ & $\underline{16.4}^\circ$ & $18.9^\circ$ & $\underline{24.5}^\circ$ & $\underline{18.5}^\circ$
  \\  %\hline
  &Ours & $\textbf{15.0}^\circ$ & $\textbf{11.1}^\circ$ & $\textbf{13.1}^\circ$ & $\textbf{12.6}^\circ$ & $\textbf{12.2}^\circ$ & $\textbf{10.2}^\circ$ & $\textbf{11.5}^\circ$ & $\textbf{17.9}^\circ$ & $21.2^\circ$ & $\textbf{16.1}^\circ$ & $\textbf{18.4}^\circ$ & $\textbf{12.9}^\circ$ & $\textbf{13.3}^\circ$ & $\textbf{19.5}^\circ$ & $\textbf{16.0}^\circ$ & $\textbf{8.6}^\circ$ & $\textbf{13.9}^\circ$ & $\textbf{13.5}^\circ$ & $\textbf{13.2}^\circ$
  \\ \midrule
  \multirow{4}{*}{\textit{present}}&\textit{Head Direction} & $17.6^\circ$ & $16.2^\circ$ & $16.9^\circ$ & $18.5^\circ$ & $25.3^\circ$ & $22.9^\circ$ & $22.3^\circ$ & $\underline{20.9}^\circ$ & $19.9^\circ$ & $18.6^\circ$ & $19.8^\circ$ & $\underline{12.4}^\circ$ & $16.8^\circ$ & $\underline{19.0}^\circ$ & $16.6^\circ$ & $16.6^\circ$ & $\underline{14.3}^\circ$ & $23.7^\circ$ & $17.7^\circ$
  \\ %\hline  
  &\textit{DGaze}\cite{hu2020dgaze} & $13.4^\circ$ & $12.1^\circ$ & $12.8^\circ$ & $\underline{10.3}^\circ$ & $\underline{10.8}^\circ$ & $\underline{8.8}^\circ$ & $\underline{9.9}^\circ$ & $22.6^\circ$ & $20.5^\circ$ & $\underline{17.3}^\circ$ & $20.2^\circ$ & $14.1^\circ$ & $16.5^\circ$ & $22.0^\circ$ & $16.3^\circ$ & $\underline{14.8}^\circ$ & $17.4^\circ$ & $24.1^\circ$ & $17.5^\circ$
  \\ %\hline
  &\textit{FixationNet}\cite{hu2021fixationnet} & $\underline{13.2}^\circ$ & $\underline{11.7}^\circ$ & $\underline{12.5}^\circ$ & $11.2^\circ$ & $11.7^\circ$ & $9.5^\circ$ & $10.6^\circ$ & $21.7^\circ$ & $\underline{19.6}^\circ$ & $17.5^\circ$ & $\underline{19.7}^\circ$ & $13.9^\circ$ & $\underline{16.3}^\circ$ & $21.8^\circ$ & $\underline{16.1}^\circ$ & $15.1^\circ$ & $17.2^\circ$ & $\underline{23.7}^\circ$ & $\underline{17.3}^\circ$
  \\  %\hline
  &Ours & $\textbf{10.7}^\circ$ & $\textbf{9.4}^\circ$ & $\textbf{10.1}^\circ$ & $\textbf{9.5}^\circ$ & $\textbf{9.8}^\circ$ & $\textbf{8.1}^\circ$ & $\textbf{9.0}^\circ$ & $\textbf{15.9}^\circ$ & $\textbf{17.3}^\circ$ & $\textbf{15.9}^\circ$ & $\textbf{16.3}^\circ$ & $\textbf{12.1}^\circ$ & $\textbf{13.5}^\circ$ & $\textbf{16.7}^\circ$ & $\textbf{14.2}^\circ$ & $\textbf{9.7}^\circ$ & $\textbf{12.0}^\circ$ & $\textbf{13.0}^\circ$ & $\textbf{13.0}^\circ$
  \\ \midrule  
  \multirow{4}{*}{\textit{future}}&\textit{Head Direction} & $17.6^\circ$ & $16.2^\circ$ & $16.9^\circ$ & $18.5^\circ$ & $25.3^\circ$ & $22.9^\circ$ & $22.3^\circ$ & $\underline{20.9}^\circ$ & $19.9^\circ$ & $18.6^\circ$ & $19.8^\circ$ & $\underline{12.4}^\circ$ & $16.8^\circ$ & $\underline{19.0}^\circ$ & $16.6^\circ$ & $16.6^\circ$ & $\underline{14.3}^\circ$ & $23.7^\circ$ & $17.7^\circ$
  \\ %\hline
  &\textit{DGaze}\cite{hu2020dgaze} & $13.4^\circ$ & $12.1^\circ$ & $12.8^\circ$ & $\underline{10.3}^\circ$ & $\underline{10.8}^\circ$ & $\underline{8.8}^\circ$ & $\underline{9.9}^\circ$ & $22.6^\circ$ & $20.5^\circ$ & $\underline{17.3}^\circ$ & $20.2^\circ$ & $14.1^\circ$ & $16.5^\circ$ & $22.0^\circ$ & $16.3^\circ$ & $\underline{14.8}^\circ$ & $17.4^\circ$ & $24.1^\circ$ & $17.5^\circ$
  \\ %\hline
  &\textit{FixationNet}\cite{hu2021fixationnet} & $\underline{13.2}^\circ$ & $\underline{11.7}^\circ$ & $\underline{12.5}^\circ$ & $11.2^\circ$ & $11.7^\circ$ & $9.5^\circ$ & $10.6^\circ$ & $21.7^\circ$ & $\underline{19.6}^\circ$ & $17.5^\circ$ & $\underline{19.7}^\circ$ & $13.9^\circ$ & $\underline{16.3}^\circ$ & $21.8^\circ$ & $\underline{16.1}^\circ$ & $15.1^\circ$ & $17.2^\circ$ & $\underline{23.7}^\circ$ & $\underline{17.3}^\circ$
  \\  %\hline
  &Ours & $\textbf{10.1}^\circ$ & $\textbf{8.8}^\circ$ & $\textbf{9.5}^\circ$ & $\textbf{9.7}^\circ$ & $\textbf{9.3}^\circ$ & $\textbf{7.9}^\circ$ & $\textbf{8.9}^\circ$ & $\textbf{15.4}^\circ$ & $\textbf{16.2}^\circ$ & $\textbf{14.8}^\circ$ & $\textbf{15.5}^\circ$ & $\textbf{11.1}^\circ$ & $\textbf{13.2}^\circ$ & $\textbf{15.8}^\circ$ & $\textbf{14.5}^\circ$ & $\textbf{9.2}^\circ$ & $\textbf{11.9}^\circ$ & $\textbf{13.9}^\circ$ & $\textbf{12.9}^\circ$
  \\ 
        \bottomrule
	\end{tabular}}
\end{table*}

\begin{figure}[htbp]
    \centering
    \includegraphics[width=0.49\textwidth]{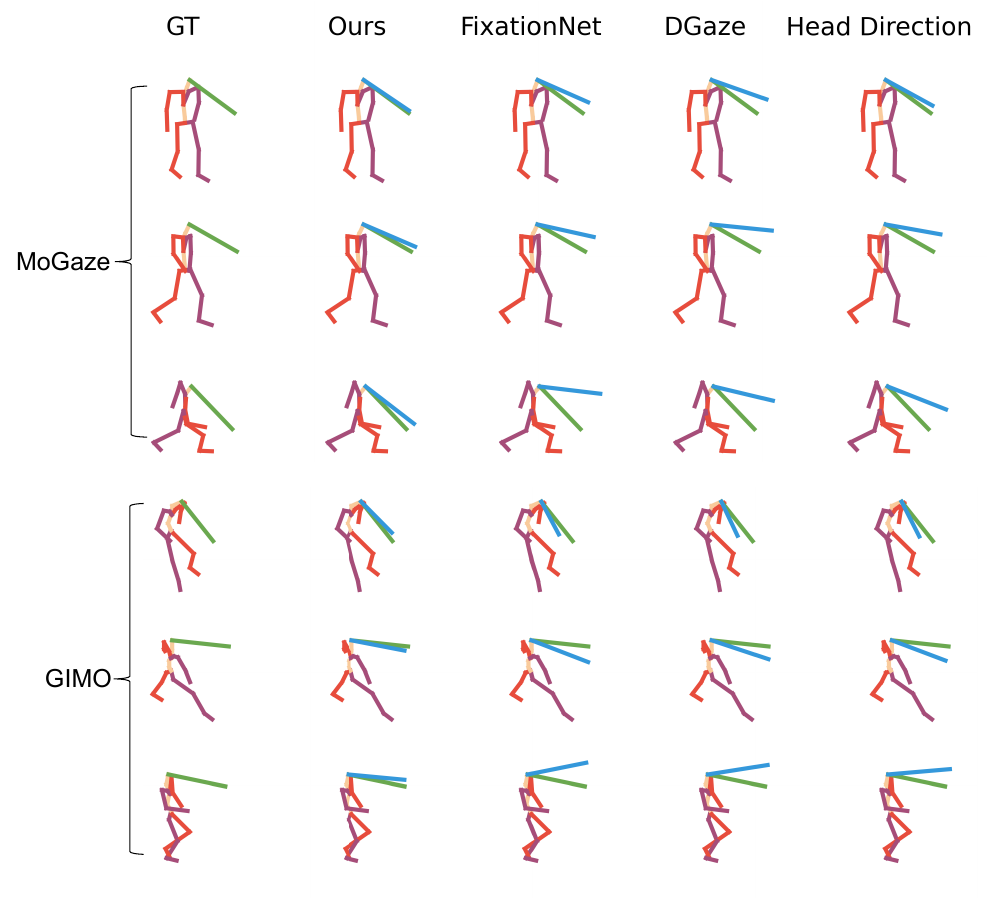}
    \caption{Results of different methods for generating eye gaze from present poses on the MoGaze and GIMO datasets. The green line indicates the ground truth while the blue line represents the predicted eye gaze.}\label{fig:results}
\end{figure}

\paragraph{Generating Gaze from Present Poses}
\autoref{tab:results}\textit{-present} summarises the mean angular errors of different methods for generating eye gaze from present body poses on the MoGaze, ADT, GIMO, and EgoBody datasets.
We can see that in terms of both individual actions and all the actions in a dataset, our method achieves better performances than the state of the art.
Specifically, our method achieves an average improvement of $19.2\%$ ($10.1^\circ$ \textit{vs.} $12.5^\circ$) on MoGaze, $9.1\%$ ($9.0^\circ$ \textit{vs.} $9.9^\circ$) on ADT, $17.3\%$ ($16.3^\circ$ \textit{vs.} $19.7^\circ$) on GIMO, and $24.9\%$ ($13.0^\circ$ \textit{vs.} $17.3^\circ$) on EgoBody.
A paired Wilcoxon signed-rank test was performed to compare the mean angular error of our method with that of the state-of-the-art methods and the results indicated that the differences between our method and the state of the art are statistically significant ($p<0.01$) on all the four datasets.
In addition, we find that our method achieves the highest improvement ($24.9\%$) on the EgoBody dataset, demonstrating its superiority for human-human interaction activities.
\autoref{fig:results} illustrates the prediction results of different methods on the MoGaze and GIMO datasets.
We can see that the eye gaze directions predicted by our method are more close to the ground truth compared with that generated by other methods (see supplementary video for more results).
These results demonstrate that our method is able to generate eye gaze from present body poses in various daily activities.

\paragraph{Generating Gaze from Future Poses}
\autoref{tab:results}\textit{-future} summarises the mean angular errors of different methods for generating eye gaze from future body poses on the MoGaze, ADT, GIMO, and EgoBody datasets.
We can see that our method outperforms prior methods in both individual actions and all the actions in a dataset, achieving an average improvement of $24.0\%$ ($9.5^\circ$ \textit{vs.} $12.5^\circ$) on MoGaze, $10.1\%$ ($8.9^\circ$ \textit{vs.} $9.9^\circ$) on ADT, $21.3\%$ ($15.5^\circ$ \textit{vs.} $19.7^\circ$) on GIMO, and $25.4\%$ ($12.9^\circ$ \textit{vs.} $17.3^\circ$) on EgoBody.
A paired Wilcoxon signed-rank test was conducted and the results validated that the differences between our method and the state of the art are statistically significant ($p<0.01$) in all the four datasets.
The above results demonstrate that our method has high performance in generating eye gaze from future body poses in both human-object and human-human interaction activities.

\paragraph{Summary}
The above results demonstrate that our method significantly outperforms the state-of-the-art methods for three different eye gaze generation tasks.
Furthermore, we find that in human-object interaction activities, our method achieves significantly higher accuracies using future body poses than using past body poses: $9.5^\circ$ \textit{vs.} $13.1^\circ$ on MoGaze, $8.9^\circ$ \textit{vs.} $11.5^\circ$ on ADT, and $15.5^\circ$ \textit{vs.} $18.4^\circ$ on GIMO.
These results correspond with our analysis in \autoref{sec:gaze_body_motion} that eye gaze has highest correlations with body motions in the near future and suggest that in offline applications where future pose is available, e.g. virtual content design and optimisation~\cite{sitzmann2018saliency,alghofaili2019optimizing}, using future body poses could generate more accurate eye gaze directions.
\hl{In human-human interaction activities, we find that our method using past, present, or future body poses achieve similar performances: $13.2^\circ$ \textit{vs.} $13.0^\circ$ \textit{vs.} $12.9^\circ$ on EgoBody.
This is because the eye-body correlations in human-human interactions are very close at $-500$ $ms$ (past), $0$ $ms$ (present) or $500$ $ms$ (future), as illustrated in \autoref{fig:gaze_two_body_motion}b.
%This corresponds with our findings in \autoref{sec:gaze_two_body_motion} that there is little or no time delay between body motions and eye gaze in human-human interactions.
}

\paragraph{Model Size and Time Costs}
Our model has smaller size than the state-of-the-art methods, containing only $0.17M$ trainable parameters while \textit{DGaze} has $0.27M$ parameters and \textit{FixationNet} contains $0.37M$ parameters.
We implemented our model on an NVIDIA Tesla V100 SXM2 32GB GPU with an Intel(R) Xeon(R) Platinum 8260 CPU @ 2.40GHz and calculated its time costs.
We find that our model requires only $35$ ms to train per batch and $3$ ms to test per batch, suggesting that our model is fast enough for practical usage.
%\andreas{any way you could compare this to other models, in particular the baselines? otherwise it just stands on its own and it's difficult to judge how much smaller it is.}

\subsection{Ablation Study}
We finally performed an ablation study to evaluate the effectiveness of each component used in our model.
Specifically, we respectively evaluated different ablated versions of our model that did not contain the \textit{DCT}, \textit{spatial GCN}, \textit{temporal GCN}, the body poses of the person, the body poses of the interaction partner, and the head directions under three generation settings.
\autoref{tab:ablation} summarises the mean angular errors of different ablated versions of our method on the ADT, GIMO, and EgoBody datasets.
We can see that our method consistently outperforms the ablated methods under different generation settings and the results are statistically significant (paired Wilcoxon signed-rank test, $p<0.01$), thus underlining the effectiveness of our model architecture.
%One exception is that our method cannot outperform the \textit{w/o DCT} or \textit{w/o T-GCN} versions in some cases.
%This is probably because both \textit{DCT} and \textit{temporal GCN} are used to extract temporal features from the sequence of body poses and are complementary to each other to some extent. 
%As a consequence, removing one of them does not influence the overall performance in some cases.
In addition, we find that in human-human interactions removing the body poses of the person or the body poses of the interaction partner significantly degrades the performance of our model: EgoBody\textit{-past} from $13.2^\circ$ to $20.6^\circ$ or $18.6^\circ$, EgoBody\textit{-present} from $13.0^\circ$ to $18.1^\circ$ or $17.9^\circ$, EgoBody\textit{-future} from $12.9^\circ$ to $17.1^\circ$ or $17.9^\circ$.
These results demonstrate the effectiveness of the two body poses on generating eye gaze in human-human interaction activities.

\begin{table}[t] 
	\centering
	% set the caption of the table.
	\caption{Mean angular errors of different ablated versions of our method on the ADT, GIMO, and EgoBody datasets. \textit{Pose} refers to the body pose of the person while \textit{Pose\_I} means the body pose of the interaction partner.} \label{tab:ablation} 
        \resizebox{0.5\textwidth}{!}{
	\begin{tabular}{ccccccccc}
		\toprule
		      && Ours & w/o \textit{DCT} & w/o \textit{S-GCN} & w/o \textit{T-GCN} & w/o \textit{Pose} & w/o \textit{Pose\_I} & w/o \textit{Head} \\ \hline
   	        \multirow{3}{*}{ADT} &\textit{past} & $\textbf{11.5}^\circ$ & $11.7^\circ$ & $11.8^\circ$ & $11.9^\circ$ & $12.2^\circ$ & - & $18.2^\circ$
            \\ %\hline
                &\textit{present} & $\textbf{9.0}^\circ$ & $9.1^\circ$ & $9.4^\circ$ & $9.1^\circ$ & $9.5^\circ$ & - & $17.7^\circ$ 
                \\ %\hline
   	        &\textit{future} & $\textbf{8.9}^\circ$ & $9.1^\circ$ & $9.3^\circ$ & $9.1^\circ$ & $9.3^\circ$ & - & $16.4^\circ$             				
            \\\midrule 
   	    \multirow{3}{*}{GIMO} &\textit{past} & $\textbf{18.4}^\circ$ & $19.0^\circ$ & $19.3^\circ$ & $19.1^\circ$ & $21.2^\circ$ & - & $22.1^\circ$      \\ %\hline            
            &\textit{present} & $\textbf{16.3}^\circ$ & $17.3^\circ$ & $18.1^\circ$ & $17.3^\circ$ & $20.8^\circ$ & - & $20.9^\circ$ \\ %\hline
   	    &\textit{future} & $\textbf{15.5}^\circ$ & $16.6^\circ$ & $18.1^\circ$ & $16.7^\circ$ & $20.8^\circ$ & - & $18.8^\circ$ \\\midrule
   	        \multirow{3}{*}{EgoBody} &\textit{past} & $\textbf{13.2}^\circ$ & $13.5^\circ$ & $13.4^\circ$ & $13.4^\circ$ & $20.6^\circ$ & $18.6^\circ$ & $15.1^\circ$            
            \\ %\hline
                &\textit{present} & $\textbf{13.0}^\circ$ & $13.1^\circ$ & $14.3^\circ$ & $13.3^\circ$ & $18.1^\circ$ & $17.9^\circ$ & $14.5^\circ$
                \\ %\hline
   	        &\textit{future} & $\textbf{12.9}^\circ$ & $13.7^\circ$ & $14.8^\circ$ & $13.5^\circ$ & $17.1^\circ$ & $17.9^\circ$ & $15.1^\circ$
            \\
        \bottomrule
	\end{tabular}}
\end{table}

%% file: sections/application.tex
\section{Eye-based Activity Recognition}\label{sec:application}

%\andreas{call it eye-based activity recognition everywhere -- that's the established term}
Activity recognition is a popular task in the area of human-centred computing and has great relevance for many VR/AR scenarios including adaptive virtual environment design~\cite{hadnett2019effect}, low-latency predictive interfaces~\cite{david2021towards, keshava2020decoding}, or human-aware intelligent systems \cite{vortmann2020attention}.
It is well-known that
%he close link between user activities and eye movements and revealed that 
user activities can be recognised directly from human eye gaze~\cite{hu2022ehtask, bulling2010eye}.
As such, eye-based activity recognition serves as a particularly relevant sample downstream task to further evaluate our gaze prediction method.
The higher the activity recognition accuracy, the better the quality of the predicted gaze sequences.
%'s effectiveness in real applications, we compared our method with other methods nn the downstream task of eye-based activity recognition.

\paragraph{Datasets} We only evaluated on the ADT and EgoBody datasets given that they provide activity labels for each recorded sequence and their sequence lengths are sufficient ($>$ 10 s) for training activity recognition methods~\cite{hu2022ehtask, coutrot2018scanpath}.
For both datasets, we used the same training and test sets as in \autoref{sec:experiments}.

\paragraph{Activity Recognition Method} We used the state-of-the-art method \textit{EHTask}~\cite{hu2022ehtask} for activity recognition in VR. \textit{EHTask} takes a sequence of eye gaze data as input, extracts eye features using a 1D CNN and a bidirectional gated recurrent unit (GRU), and finally recognises user activities from the eye features using fully-connected layers.

\paragraph{Procedure}
%Our evaluation procedure contains two stages: a training stage and a test stage.
%In the training stage,
%We trained the different gaze prediction methods on the ADT and EgoBody datasets from \textit{present} 15 frames as in \autoref{sec:experiments}.
We first trained \textit{EHTask} to recognise user activities from the ground truth eye gaze sequences using default parameters.
On the ADT dataset, we trained \textit{EHTask} to recognise three activities: \textit{decoration}, \textit{meal}, and \textit{work}.
On the EgoBody dataset, we
%first removed some activities that contain only a few recordings and then
trained \textit{EHTask} for three activities that have the most recordings, i.e. \textit{chat}, \textit{learn}, and \textit{teach}.
Since activity recognition usually requires a long sequence as input~\cite{hu2022ehtask, coutrot2018scanpath}, we used 300 frames (corresponding to 10 seconds) of ground truth eye gaze data to train \textit{EHTask} as in the original paper~\cite{hu2022ehtask}.
%\andreas{at test time you used the ground truth gaze sequences I assume?}\zhiming{I used ground truth eye gaze to train EHTask and used the eye gaze predicted from different method to test EHTask.}
At test time, we used the eye gaze sequences generated from different methods as input to \textit{EHTask} and evaluated its activity recognition performance.
Since the gaze prediction methods were trained to predict 15 frames of eye gaze while \textit{EHTask} requires an input of 300 frames, we first segmented the 300 frames of test data into 20 windows (each window contains 15 frames), then predicted eye gaze for each window, and finally concatenated the gaze predictions of all the windows and used them as input to \textit{EHTask} to recognise user activities.

\paragraph{Results} \autoref{tab:activity_recognition} shows the recognition performances of using the ground truth eye gaze and the eye gaze predicted from different methods on ADT and EgoBody.
%the ground truth eye gaze and different methods on ADT and EgoBody when using predicted gaze sequences.
It can be seen in the table that our method achieves higher recognition accuracies than other methods on both ADT ($70.0\%$ vs. $67.3\%$) and EgoBody ($60.1\%$ vs. $58.2\%$).
Using a paired Wilcoxon signed-rank test
%to compare the performances of our method with other methods and validated
we confirmed that these differences were statistically significant ($p<0.01$).
\hl{We also notice that our method can achieve comparable performance with the ground truth eye gaze in terms of recognition accuracy ($70.0\%$ vs. $74.7\%$ on ADT, and $60.1\%$ vs. $62.1\%$ on EgoBody).}
These results illustrate the benefit of our method when using the predicted gaze sequences for downstream tasks with high relevance for VR/AR applications, such as
%significantly outperforms prior methods in the real application of
eye-based activity recognition.

\begin{table}[t] 
	\centering
	% set the caption of the table.
	\caption{Eye-based activity recognition accuracies of using ground truth eye gaze and the eye gaze predicted from different methods on ADT and EgoBody. Best results are in bold while the second best are underlined.} \label{tab:activity_recognition} 
        \resizebox{0.45\textwidth}{!}{
	\begin{tabular}{ccccccc}
		\toprule
		      &GT & Ours &\textit{DGaze}\cite{hu2020dgaze} &\textit{FixationNet}\cite{hu2021fixationnet} &\textit{Head Direction} &\textit{Chance}\\ \hline
   	        ADT &\textbf{74.7\%}
 &\underline{70.0\%} &67.3\% &66.8\% &40.9\% &33.3\%\\ %\hline
                EgoBody &\textbf{62.1\%}
 &\underline{60.1\%} &52.3\% &58.2\% &50.3\% &33.3\%\\
        \bottomrule
	\end{tabular}}
\end{table}

%% file: sections/discussion.tex
\section{Discussion} \label{sec:discussion}

In this work we have made an important step towards understanding the correlation between eye and full-body movements in daily activities and generating eye gaze from full-body poses.

%Our analyses and the new method advance research in the fields of virtual reality, human-computer interaction, and human-centred computing in several ways.

%\andreas{the following paragraphs are not wrong but they fall a bit short in making clear what the core/novel contributions/findings are. The first two also already talk about applications - and then you have a dedicated paragraph on applications. This is not ideal IMHO. Think about what is the core take away message(s) for each part (coordination analysis, pose-based gaze generation, applications)}

\paragraph{Eye-body Coordination}
Our analyses on eye-body coordination in daily activities revealed novel insights for understanding human eye and body movements that we also used to guide the design of our eye-body coordination model.
%Specifically, in the analysis of correlation between eye gaze and body orientations, we find that eye gaze has strong correlations with the orientations of different body joints, especially with head directions (\autoref{tab:gaze_body_orientation}).
%Inspired by this, our model extracts body orientation features from head directions for eye gaze prediction (\autoref{sec:body_orientation_feature}).
Specifically, in the analysis of correlation between gaze and head directions, we found that there exists little or no time delay between head and eye movements (see \autoref{fig:gaze_head_direction}).
Therefore, in the task of generating gaze from future poses (see \autoref{sec:experimental_settings}), we used head directions at the present time to generate target eye gaze rather than using future head directions.
In the analysis of correlation between eye gaze and body motions, we found that gaze is strongly correlated with full-body motions in human-object interactions (see \autoref{sec:gaze_body_motion}) and has high correlations with the directions between two bodies in human-human interactions (see \autoref{sec:gaze_two_body_motion}).
Therefore, our model uses features from the body pose of the person to generate gaze in human-object interactions and employed the features from the body poses of the person and the interaction partner to predict gaze in human-human interactions (see \autoref{fig:method}).
We also found that eye movements precede body motions in human-object interactions (see \autoref{sec:gaze_body_motion}).
Exploiting this, we used future body poses as input to improve the performance of eye gaze generation in situations where future pose information is available (see \autoref{sec:gaze_generation_results}).
The superior performance of our model shows that our analyses are effective and are valuable for the design of future eye-body coordination models.

\paragraph{Pose-based Gaze Prediction}
Prior methods have typically generated eye gaze from head movements while we predict eye gaze from both head movements and full-body poses.
This approach proved highly effective and significantly outperforms prior methods in both human-object and human-human interaction activities for three different eye gaze generation tasks (see \autoref{tab:results}).
An ablation study confirmed that body poses are instrumental for achieving these performance improvements (see \autoref{tab:ablation}).
Taken together, these results underline the significant potential of human full-body poses for generating eye movements and thus open the promising research direction of pose-based gaze prediction.

\paragraph{Applications of Our Method}
In addition to the sample downstream task of eye-based activity recognition (see \autoref{sec:application}), we believe that our method
%is also relevant for various gaze-based VR/AR applications.
%The fact that our method outperforms the state-of-the-art methods by a large margin under all the three different generation settings (see \autoref{tab:results}) demonstrates that our method
has significant potential for numerous VR/AR applications, including dynamic event triggering~\cite{hadnett2019effect}, gaze-contingent rendering~\cite{hu2019sgaze, hu2020dgaze}, gaze-based interaction~\cite{duchowski2018gaze}, or virtual content design and optimisation~\cite{sitzmann2018saliency, alghofaili2019optimizing}.
In addition, with further advances in the development of motion capture and motion generation techniques~\cite{bhattacharya2021text2gestures,hasegawa2018evaluation}, it will become increasingly easy to obtain human full-body poses in VR and AR environments.
In the future, our approach could be integrated with such motion capture and motion generation techniques to generate eye and full-body movements simultaneously.

\paragraph{Limitations and Future Work}
Despite these advances, our work has several limitations that we plan to address in the future.
First, we focused on eye-body coordination in different activities and ignored the influence of the visual scene in front of the person.
It will be interesting to explore whether eye-body coordination during the same activity will be affected by different scene content.
If so, this will also open a new avenue for research on ways to incorporate visual scene information into the prediction and to see whether this can improve performance further.
Second, incorporating other modalities such as facial expressions and audio signals in social interactions may also boost our model's gaze prediction performance.
\hl{In addition, existing datasets that include body pose and eye gaze have all been collected in indoor environments, thus unfortunately limiting our analyses to indoor settings.
It remains to be seen whether eye-body coordination as characterised here also generalises to outdoor scenarios.
Furthermore, existing datasets only offer data for interactions between two humans, thus limiting our analyses to dyadic settings.
It will be interesting to explore the eye-body coordination during other interaction settings that are important for VR/AR applications, e.g. interactions between more humans and interactions between a human and a virtual avatar.}
Finally, generating stylistic eye gaze, e.g. eye gaze that can convey different emotions~\cite{randhavane2019eva}, is an interesting avenue for extending our work in the future.

%\paragraph{Future Work}
%Besides overcoming the above limitations, many potential avenues of future work exist.
%First, our method predicts human eye gaze directly from human head directions and full-body poses.
%It would be promising to also incorporate the scene content into our method to further improve its performance.
%Finally, we are looking forward to optimising our model in various VR/AR applications, such as gaze-based interaction and gaze-contingent rendering.

%% file: sections/conclusion.tex
\section{Conclusion}
In this work we were the first to explore the challenging task of predicting human eye gaze from full-body poses.
We first conducted a comprehensive analysis on the coordination of human eye and full-body movements in everyday activities and revealed that in human-object interactions eye gaze is strongly correlated with full-body motions while in human-human interactions a person's gaze direction has strong correlations with the directions pointing from his body to the body of the interaction partner.
Based on these insights, we proposed a novel eye-body coordination model to predict eye gaze from head directions and full-body poses that outperformed the state-of-the-art methods by a large margin for three different prediction tasks as well as the application of eye-based activity recognition.
Taken together, our work provides novel insights into eye-body coordination during daily activities and makes an important step towards pose-based gaze prediction.

%% file: main.bbl
\begin{thebibliography}{10}
\providecommand{\url}[1]{#1}
\csname url@samestyle\endcsname
\providecommand{\newblock}{\relax}
\providecommand{\bibinfo}[2]{#2}
\providecommand{\BIBentrySTDinterwordspacing}{\spaceskip=0pt\relax}
\providecommand{\BIBentryALTinterwordstretchfactor}{4}
\providecommand{\BIBentryALTinterwordspacing}{\spaceskip=\fontdimen2\font plus
\BIBentryALTinterwordstretchfactor\fontdimen3\font minus
  \fontdimen4\font\relax}
\providecommand{\BIBforeignlanguage}[2]{{%
\expandafter\ifx\csname l@#1\endcsname\relax
\typeout{** WARNING: IEEEtran.bst: No hyphenation pattern has been}%
\typeout{** loaded for the language `#1'. Using the pattern for}%
\typeout{** the default language instead.}%
\else
\language=\csname l@#1\endcsname
\fi
#2}}
\providecommand{\BIBdecl}{\relax}
\BIBdecl

\bibitem{hu2021fixationnet}
Z.~Hu, A.~Bulling, S.~Li, and G.~Wang, ``Fixationnet: forecasting eye fixations
  in task-oriented virtual environments,'' \emph{IEEE Transactions on
  Visualization and Computer Graphics}, vol.~27, no.~5, pp. 2681--2690, 2021.

\bibitem{hadnett2019effect}
J.~Hadnett-Hunter, G.~Nicolaou, E.~O'Neill, and M.~Proulx, ``The effect of task
  on visual attention in interactive virtual environments,'' \emph{ACM
  Transactions on Applied Perception}, vol.~16, no.~3, pp. 1--17, 2019.

\bibitem{patney2016towards}
A.~Patney, M.~Salvi, J.~Kim, A.~Kaplanyan, C.~Wyman, N.~Benty, D.~Luebke, and
  A.~Lefohn, ``Towards foveated rendering for gaze-tracked virtual reality,''
  \emph{ACM Transactions on Graphics}, vol.~35, no.~6, pp. 1--12, 2016.

\bibitem{sidenmark2019eyehead}
L.~Sidenmark and H.~Gellersen, ``Eye\&head: Synergetic eye and head movement
  for gaze pointing and selection,'' in \emph{Proceedings of the 2019 Annual
  ACM Symposium on User Interface Software and Technology}, 2019, pp.
  1161--1174.

\bibitem{alghofaili2019optimizing}
R.~Alghofaili, M.~S. Solah, H.~Huang, Y.~Sawahata, M.~Pomplun, and L.-F. Yu,
  ``Optimizing visual element placement via visual attention analysis,'' in
  \emph{Proceedings of the 2019 IEEE Conference on Virtual Reality and 3D User
  Interfaces}.\hskip 1em plus 0.5em minus 0.4em\relax IEEE, 2019, pp. 464--473.

\bibitem{sun2018towards}
Q.~Sun, A.~Patney, L.-Y. Wei, O.~Shapira, J.~Lu, P.~Asente, S.~Zhu, M.~McGuire,
  D.~Luebke, and A.~Kaufman, ``Towards virtual reality infinite walking:
  dynamic saccadic redirection,'' \emph{ACM Transactions on Graphics}, vol.~37,
  no.~4, pp. 1--13, 2018.

\bibitem{hu2022ehtask}
Z.~Hu, A.~Bulling, S.~Li, and G.~Wang, ``Ehtask: recognizing user tasks from
  eye and head movements in immersive virtual reality,'' \emph{IEEE
  Transactions on Visualization and Computer Graphics}, 2022.

\bibitem{canigueral2019being}
R.~Ca{\~n}igueral and A.~F. d.~C. Hamilton, ``Being watched: Effects of an
  audience on eye gaze and prosocial behaviour,'' \emph{Acta Psychologica},
  vol. 195, pp. 50--63, 2019.

\bibitem{sitzmann2018saliency}
V.~Sitzmann, A.~Serrano, A.~Pavel, M.~Agrawala, D.~Gutierrez, B.~Masia, and
  G.~Wetzstein, ``Saliency in vr: how do people explore virtual environments?''
  \emph{IEEE Transactions on Visualization and Computer Graphics}, vol.~24,
  no.~4, pp. 1633--1642, 2018.

\bibitem{hu2019sgaze}
Z.~Hu, C.~Zhang, S.~Li, G.~Wang, and D.~Manocha, ``Sgaze: a data-driven
  eye-head coordination model for realtime gaze prediction,'' \emph{IEEE
  Transactions on Visualization and Computer Graphics}, vol.~25, no.~5, pp.
  2002--2010, 2019.

\bibitem{hu2020dgaze}
Z.~Hu, S.~Li, C.~Zhang, K.~Yi, G.~Wang, and D.~Manocha, ``Dgaze: Cnn-based gaze
  prediction in dynamic scenes,'' \emph{IEEE Transactions on Visualization and
  Computer Graphics}, vol.~26, no.~5, pp. 1902--1911, 2020.

\bibitem{kratzer2020mogaze}
P.~Kratzer, S.~Bihlmaier, N.~B. Midlagajni, R.~Prakash, M.~Toussaint, and
  J.~Mainprice, ``Mogaze: A dataset of full-body motions that includes
  workspace geometry and eye-gaze,'' \emph{IEEE Robotics and Automation
  Letters}, vol.~6, no.~2, pp. 367--373, 2020.

\bibitem{pan2023aria}
X.~Pan, N.~Charron, Y.~Yang, S.~Peters, T.~Whelan, C.~Kong, O.~Parkhi,
  R.~Newcombe, and Y.~C. Ren, ``Aria digital twin: A new benchmark dataset for
  egocentric 3d machine perception,'' in \emph{Proceedings of the IEEE/CVF
  International Conference on Computer Vision}, 2023, pp. 20\,133--20\,143.

\bibitem{zheng2022gimo}
Y.~Zheng, Y.~Yang, K.~Mo, J.~Li, T.~Yu, Y.~Liu, K.~Liu, and L.~J. Guibas,
  ``Gimo: Gaze-informed human motion prediction in context,'' in
  \emph{Proceedings of the 2022 European Conference on Computer Vision}, 2022.

\bibitem{zhang2022egobody}
S.~Zhang, Q.~Ma, Y.~Zhang, Z.~Qian, M.~Pollefeys, F.~Bogo, and S.~Tang,
  ``Egobody: Human body shape, motion and social interactions from head-mounted
  devices,'' in \emph{Proceedings of the 2022 European Conference on Computer
  Vision}, 2022.

\bibitem{hasegawa2018evaluation}
D.~Hasegawa, N.~Kaneko, S.~Shirakawa, H.~Sakuta, and K.~Sumi, ``Evaluation of
  speech-to-gesture generation using bi-directional lstm network,'' in
  \emph{Proceedings of the 2018 International Conference on Intelligent Virtual
  Agents}, 2018, pp. 79--86.

\bibitem{kucherenko2019analyzing}
T.~Kucherenko, D.~Hasegawa, G.~E. Henter, N.~Kaneko, and H.~Kjellstr{\"o}m,
  ``Analyzing input and output representations for speech-driven gesture
  generation,'' in \emph{Proceedings of the 2019 ACM International Conference
  on Intelligent Virtual Agents}, 2019, pp. 97--104.

\bibitem{yoon2019robots}
Y.~Yoon, W.-R. Ko, M.~Jang, J.~Lee, J.~Kim, and G.~Lee, ``Robots learn social
  skills: End-to-end learning of co-speech gesture generation for humanoid
  robots,'' in \emph{Proceedings of the 2019 International Conference on
  Robotics and Automation}.\hskip 1em plus 0.5em minus 0.4em\relax IEEE, 2019,
  pp. 4303--4309.

\bibitem{bhattacharya2021text2gestures}
U.~Bhattacharya, N.~Rewkowski, A.~Banerjee, P.~Guhan, A.~Bera, and D.~Manocha,
  ``Text2gestures: A transformer-based network for generating emotive body
  gestures for virtual agents,'' in \emph{Proceedings of the 2021 IEEE Virtual
  Reality and 3D User Interfaces}.\hskip 1em plus 0.5em minus 0.4em\relax IEEE,
  2021, pp. 1--10.

\bibitem{ye2020choreonet}
Z.~Ye, H.~Wu, J.~Jia, Y.~Bu, W.~Chen, F.~Meng, and Y.~Wang, ``Choreonet:
  Towards music to dance synthesis with choreographic action unit,'' in
  \emph{Proceedings of the 2020 ACM International Conference on Multimedia},
  2020, pp. 744--752.

\bibitem{lee2019dancing}
H.-Y. Lee, X.~Yang, M.-Y. Liu, T.-C. Wang, Y.-D. Lu, M.-H. Yang, and J.~Kautz,
  ``Dancing to music,'' \emph{Advances in Neural Information Processing
  Systems}, vol.~32, 2019.

\bibitem{higuch2016can}
K.~Higuch, R.~Yonetani, and Y.~Sato, ``Can eye help you? effects of visualizing
  eye fixations on remote collaboration scenarios for physical tasks,'' in
  \emph{Proceedings of the 2016 CHI Conference on Human Factors in Computing
  Systems}, 2016, pp. 5180--5190.

\bibitem{duarte2018action}
N.~F. Duarte, M.~Rakovi{\'c}, J.~Tasevski, M.~I. Coco, A.~Billard, and
  J.~Santos-Victor, ``Action anticipation: Reading the intentions of humans and
  robots,'' \emph{IEEE Robotics and Automation Letters}, vol.~3, no.~4, pp.
  4132--4139, 2018.

\bibitem{duchowski2018gaze}
A.~T. Duchowski, ``Gaze-based interaction: a 30 year retrospective,''
  \emph{Computers and Graphics}, vol.~73, pp. 59--69, 2018.

\bibitem{jiao23supreyes}
C.~Jiao, Z.~Hu, M.~B{\^a}ce, and A.~Bulling, ``Supreyes: Super resolution for
  eyes using implicit neural representation learning,'' in \emph{Proceedings of
  the 2023 Annual ACM Symposium on User Interface Software and Technology},
  2023, pp. 1--13.

\bibitem{stahl1999amplitude}
J.~S. Stahl, ``Amplitude of human head movements associated with horizontal
  saccades,'' \emph{Experimental Brain Research}, vol. 126, no.~1, pp. 41--54,
  1999.

\bibitem{fang2015eye}
Y.~Fang, R.~Nakashima, K.~Matsumiya, I.~Kuriki, and S.~Shioiri, ``Eye-head
  coordination for visual cognitive processing,'' \emph{PloS One}, vol.~10,
  no.~3, p. e0121035, 2015.

\bibitem{kothari2020gaze}
R.~Kothari, Z.~Yang, C.~Kanan, R.~Bailey, J.~B. Pelz, and G.~J. Diaz,
  ``Gaze-in-wild: a dataset for studying eye and head coordination in everyday
  activities,'' \emph{Scientific Reports}, vol.~10, no.~1, pp. 1--18, 2020.

\bibitem{sidenmark2019eye}
L.~Sidenmark and H.~Gellersen, ``Eye, head and torso coordination during gaze
  shifts in virtual reality,'' \emph{ACM Transactions on Computer-Human
  Interaction}, vol.~27, no.~1, pp. 1--40, 2019.

\bibitem{batmaz2020touch}
A.~U. Batmaz, A.~K. Mutasim, M.~Malekmakan, E.~Sadr, and W.~Stuerzlinger,
  ``Touch the wall: Comparison of virtual and augmented reality with
  conventional 2d screen eye-hand coordination training systems,'' in
  \emph{Proceedings of the 2020 IEEE Conference on Virtual Reality and 3D User
  Interfaces}.\hskip 1em plus 0.5em minus 0.4em\relax IEEE, 2020, pp. 184--193.

\bibitem{emery2021openneeds}
K.~J. Emery, M.~Zannoli, J.~Warren, L.~Xiao, and S.~S. Talathi, ``Openneeds: A
  dataset of gaze, head, hand, and scene signals during exploration in
  open-ended vr environments,'' in \emph{Proceedings of the 2021 ACM Symposium
  on Eye Tracking Research and Applications}, 2021, pp. 1--7.

\bibitem{randhavane2019eva}
T.~Randhavane, A.~Bera, K.~Kapsaskis, R.~Sheth, K.~Gray, and D.~Manocha, ``Eva:
  Generating emotional behavior of virtual agents using expressive features of
  gait and gaze,'' in \emph{Proceedings of the 2019 ACM Symposium on Applied
  Perception}, 2019, pp. 1--10.

\bibitem{itti1998model}
L.~Itti, C.~Koch, and E.~Niebur, ``A model of saliency-based visual attention
  for rapid scene analysis,'' \emph{IEEE Transactions on Pattern Analysis and
  Machine Intelligence}, vol.~20, no.~11, pp. 1254--1259, 1998.

\bibitem{cheng2015global}
M.-M. Cheng, N.~J. Mitra, X.~Huang, P.~H. Torr, and S.-M. Hu, ``Global contrast
  based salient region detection,'' \emph{IEEE Transactions on Pattern Analysis
  and Machine Intelligence}, vol.~37, no.~3, pp. 569--582, 2015.

\bibitem{borji2012probabilistic}
A.~Borji, D.~N. Sihite, and L.~Itti, ``Probabilistic learning of task-specific
  visual attention,'' in \emph{Proceedings of the 2012 IEEE Conference on
  Computer Vision and Pattern Recognition}.\hskip 1em plus 0.5em minus
  0.4em\relax IEEE, 2012, pp. 470--477.

\bibitem{koulieris2016gaze}
G.~A. Koulieris, G.~Drettakis, D.~Cunningham, and K.~Mania, ``Gaze prediction
  using machine learning for dynamic stereo manipulation in games,'' in
  \emph{Proceedings of the 2016 IEEE Virtual Reality}.\hskip 1em plus 0.5em
  minus 0.4em\relax IEEE, 2016, pp. 113--120.

\bibitem{nakashima2015saliency}
R.~Nakashima, Y.~Fang, Y.~Hatori, A.~Hiratani, K.~Matsumiya, I.~Kuriki, and
  S.~Shioiri, ``Saliency-based gaze prediction based on head direction,''
  \emph{Vision Research}, vol. 117, pp. 59--66, 2015.

\bibitem{hu2021eye}
Z.~Hu, ``Eye fixation forecasting in task-oriented virtual reality,'' in
  \emph{Proceedings of the 2021 IEEE Conference on Virtual Reality and 3D User
  Interfaces Abstracts and Workshops}.\hskip 1em plus 0.5em minus 0.4em\relax
  IEEE, 2021, pp. 707--708.

\bibitem{ma2022progressively}
T.~Ma, Y.~Nie, C.~Long, Q.~Zhang, and G.~Li, ``Progressively generating better
  initial guesses towards next stages for high-quality human motion
  prediction,'' in \emph{Proceedings of the 2022 IEEE Conference on Computer
  Vision and Pattern Recognition}, 2022, pp. 6437--6446.

\bibitem{mao2020history}
W.~Mao, M.~Liu, and M.~Salzmann, ``History repeats itself: Human motion
  prediction via motion attention,'' in \emph{Proceedings of the 2020 European
  Conference on Computer Vision}.\hskip 1em plus 0.5em minus 0.4em\relax
  Springer, 2020, pp. 474--489.

\bibitem{el2009dynamic}
M.~S. El-Nasr, A.~Vasilakos, C.~Rao, and J.~Zupko, ``Dynamic intelligent
  lighting for directing visual attention in interactive 3d scenes,''
  \emph{IEEE Transactions on Computational Intelligence and AI in Games},
  vol.~1, no.~2, pp. 145--153, 2009.

\bibitem{steil2018forecasting}
J.~Steil, P.~M{\"{u}}ller, Y.~Sugano, and A.~Bulling, ``Forecasting user
  attention during everyday mobile interactions using device-integrated and
  wearable sensors,'' in \emph{Proceedings of the 2018 ACM International
  Conference on Human-Computer Interaction with Mobile Devices and Services},
  2018, pp. 1:1--1:13.

\bibitem{mueller2020anticipating}
P.~Müller, E.~Sood, and A.~Bulling, ``Anticipating averted gaze in dyadic
  interactions,'' in \emph{Proceedings of the 2020 ACM International Symposium
  on Eye Tracking Research and Applications}, 2020, pp. 1--10.

\bibitem{majaranta14_apc}
P.~Majaranta and A.~Bulling, \emph{Eye Tracking and Eye-Based Human-Computer
  Interaction}.\hskip 1em plus 0.5em minus 0.4em\relax Springer Publishing
  London, 2014, pp. 39--65.

\bibitem{david2021towards}
B.~David-John, C.~E. Peacock, T.~Zhang, T.~S. Murdison, H.~Benko, and T.~R.
  Jonker, ``Towards gaze-based prediction of the intent to interact in virtual
  reality,'' in \emph{Proceedings of the 2021 ACM Symposium on Eye Tracking
  Research and Applications}, 2021, pp. 1--7.

\bibitem{keshava2020decoding}
A.~Keshava, A.~Aumeistere, K.~Izdebski, and P.~Konig, ``Decoding task from
  oculomotor behavior in virtual reality,'' in \emph{Proceedings of the 2020
  ACM Symposium on Eye Tracking Research and Applications}, 2020, pp. 1--5.

\bibitem{vortmann2020attention}
L.-M. Vortmann and F.~Putze, ``Attention-aware brain computer interface to
  avoid distractions in augmented reality,'' in \emph{Extended Abstracts of the
  2020 CHI Conference on Human Factors in Computing Systems}, 2020, pp. 1--8.

\bibitem{bulling2010eye}
A.~Bulling, J.~A. Ward, H.~Gellersen, and G.~Troster, ``Eye movement analysis
  for activity recognition using electrooculography,'' \emph{IEEE Transactions
  on Pattern Analysis and Machine Intelligence}, vol.~33, no.~4, pp. 741--753,
  2010.

\bibitem{coutrot2018scanpath}
A.~Coutrot, J.~H. Hsiao, and A.~B. Chan, ``Scanpath modeling and classification
  with hidden markov models,'' \emph{Behavior Research Methods}, vol.~50,
  no.~1, pp. 362--379, 2018.

\end{thebibliography}
